%% file: 00-main.tex
\documentclass[a4paper,fleqn]{cas-dc}

\usepackage[numbers]{natbib}
\usepackage{graphicx}
\usepackage{subcaption}
\usepackage{comment}
\usepackage{orcidlink}
\usepackage{tabularx}
\usepackage{xcolor}
\usepackage{multirow}

\newcommand{\hl}[1]{{\color{black}#1}}

\usepackage{pdflscape}
\usepackage{afterpage}

\AtBeginDocument{%
  \providecommand\BibTeX{{%
    \normalfont B\kern-0.5em{\scshape i\kern-0.25em b}\kern-0.8em\TeX}}}

\def\tsc#1{\csdef{#1}{\textsc{\lowercase{#1}}\xspace}}
\tsc{WGM}
\tsc{QE}
\tsc{EP}
\tsc{PMS}
\tsc{BEC}
\tsc{DE}

\usepackage{tcolorbox}
\usepackage{fontawesome5}
\usepackage{float}
\begin{document}

\shorttitle{A Thorough Assessment of the Non-IID Data Impact in FL}

\shortauthors{Jimenez et al.}

\title [mode = title]{A Thorough Assessment of the Non-IID Data Impact in Federated Learning}                      

%

\author[1]{Daniel M. Jimenez-Gutierrez \orcidlink{0000-0002-0305-814X}}
\ead{jimenezgutierrez@diag.uniroma1.it}
\author[1]{Mehrdad Hassanzadeh \orcidlink{0000-0002-9028-104X}}
\ead{hassanzadeh.1961575@studenti.uniroma1.it}
\author[1]{Aris Anagnostopoulos \orcidlink{0000-0001-9183-7911}}
\ead{aris@diag.uniroma1.it}
\author[2]{Ioannis Chatzigiannakis \orcidlink{0000-0001-8955-9270}}
\ead{ichatz@diag.uniroma1.it}
\author[1]{Andrea Vitaletti \orcidlink{0000-0003-1074-5068}}
\ead{vitaletti@diag.uniroma1.it}

\affiliation[1]{organization={Dept. of Computer, Control and Management Engineering, Sapienza University of Rome},
    city={Rome},
    country={Italy}}

\begin{abstract}
Federated learning (FL) allows collaborative machine learning (ML) model training among decentralized clients' information, ensuring data privacy. The decentralized nature of FL deals with non-independent and identically distributed (non-IID) data. This open problem has notable consequences, such as decreased model performance and longer convergence times. Despite its importance, experimental studies systematically addressing all types of data heterogeneity (a.k.a. non-IIDness) remain scarce. This paper aims to fill this gap by assessing and quantifying the non-IID effect through an empirical analysis. We use the Hellinger Distance (\textsf{HD}) to measure differences in distribution among clients. Our study benchmarks five state-of-the-art strategies for handling non-IID data, including label, feature, quantity, and spatiotemporal skews, under realistic and controlled conditions. This is the first comprehensive analysis of the spatiotemporal skew effect in FL. Our findings highlight the significant impact of label and spatiotemporal skew non-IID types on FL model performance, with notable performance drops occurring at specific \textsf{HD} thresholds. The FL performance is also heavily affected, mainly when the non-IIDness is extreme. Thus, we provide recommendations for FL research to tackle data heterogeneity effectively. Our work represents the most extensive examination of non-IIDness in FL, offering a robust foundation for future research.
\end{abstract}

\begin{highlights}

\item Label skew and spatiotemporal skew have the most significant impact on the model's performance.

\item The drop in the model's performance for label skew appears in a double threshold. A notable performance decline is immediately evident when the Hellinger Distance exceeds 0.5 and 0.75.

\item Feature skew does not alter the model's performance nor the convergence point.

\item The quantity skew in the client's data does not affect the model's performance.

\item The higher the non-IIDness level in time and space (spatiotemporal skew), the worse the model’s performance.

\end{highlights}


\maketitle

\section{Introduction}
\label{sec:intro}
\input{01-section-intro}

\section{Related Work}
\label{sec:related-work}
\input{02-section-related-work}

\section{Background}
\label{sec:background}
\input{03-section-background}

\section{Experimentation Setup}
\label{sec:experiments}
\input{04-section-experiments}


\section{Label Skew Results}
\label{sec:label-results}
\input{05-section-label-results}

\section{Feature Skew Results}
\label{sec:feat-results}
\input{06-section-feature-results}

\section{Quantity Skew Results}
\label{sec:quant-results}
\input{07-section-quantity-results}

\section{Spatiotemporal Skew Results}
\label{sec:spatemp-results}
\input{08-section-spatemp-results}

\section{General Results}
\label{sec:general-results}
\input{09-section-general-results}

\section{Design Insights and Opportunities}
\label{sec:future-work}
\input{10-section-future-work}

\section{Conclusions}
\label{sec:conclusion}
\input{11-section-conclusion}

\section{Acknowledgments}
To be included after acceptance.

\bibliographystyle{plain}
\bibliography{references_new}

\printcredits

\end{document}

%% file: 01-section-intro.tex
In today's digital age, the interaction of machine learning (ML) and
healthcare or financial data holds immense promise for improving disease
diagnosis~\cite{rahman2024machine} and combating financial
crimes~\cite{pattnaik2024applications}. \hl{These advancements have traditionally relied on centralized learning (CL), such as Machine Learning as a Service (MLaaS) platforms—including AWS, Azure, and Google Cloud~\cite{neptunemlaas}—where raw data is aggregated on a central server for model training.}. However, it
raises critical questions about data privacy when dealing
with sensitive information from hospitals or banks. 
In this context, trusted research environments emerge as a mechanism for balancing ML research and protecting individual privacy~\cite{graham2023trust,zhang2022privacy}.

Federated learning (FL)~\cite{mcmahan2017communication} has emerged as
a transformative approach for training ML models across decentralized
data sources, preserving data privacy and security.
This paradigm is particularly beneficial in cross-silo settings, where
entities such as hospitals, banks, and other organizations collaborate
without sharing sensitive data. 
However, a significant challenge inherent in FL is the variation in data
distributions across clients, referred to as non-IID
(Non-Independent and Identically Distributed) data. This 
non-IID data (i.e., non-IIDness, data heterogeneity) hinders model performance and
convergence during training~\cite{wang2023batch,mendieta2022local}. Such \hl{non-IID data} is classified into four categories: label, feature, quantity, and spatiotemporal skew~\cite{types_noniid2}.

\emph{Spatiotemporal skew} presents unique challenges that are particularly critical yet underexplored in FL research. This skew occurs when data distributions vary across both geographical locations (spatial) and periods (temporal)~\cite{criado2022non}. Such skew fundamentally differs from the label, feature, or quantity skew by introducing dynamic variations that standard FL aggregation algorithms often fail to address~\cite{babendererde2025jointly}. Understanding spatiotemporal skew is crucial because it directly impacts the model's ability to generalize across diverse real-world environments while maintaining temporal relevance, making it a key frontier for robust FL systems.

Furthermore, diagnosing and quantifying the level of \hl{non-IID data} in FL is a significant challenge, as emphasized by Pei et al.~\cite{pei2024review} and Li et al.~\cite{li2020federated}, who identify critical research directions in this domain. Numerous studies have introduced metrics to quantify the level of \hl{non-IID data} in FL~\cite{jimenez2024fedartml,elhussein2024universal,novikova2024evaluation,wang2021accelerated}, with the Hellinger Distance (\textsf{HD})~\cite{hd_properties} emerging as one of the most reliable options. \hl{HD = 0.0 corresponds to fully IID data, while higher values (e.g., 0.25, 0.5, 0.75, 0.9) represent increasing degrees of non-IID data, with HD = 0.9 approaching the most non-IID scenario considered in our study.} HD offers a fine-grained measurement of distribution differences, achieving values close to 1 under extreme non-IID conditions, unlike the Jensen-Shannon Distance (JSD), which tends to plateau at lower levels. Furthermore, HD is versatile and applicable across various types of skews.

\paragraph{\textbf{Motivation.}} 
Recent advances in FL research have significantly advanced our understanding of \hl{non-IID data} challenges, with notable progress in addressing isolated aspects of heterogeneity such as label skew~\cite{vahidian2023rethinking,wong2023empirical,mora2022federated}. Based on this foundation, there is now a timely opportunity to unify these insights through comprehensive empirical benchmarks that span the full spectrum of non-IID skews. While pioneering theoretical frameworks~\cite{lu2024federated,pei2024review} have established critical mitigation principles, translating these into practice requires systematic quantification of how diverse \hl{non-IID data} types—from feature skew to spatiotemporal drift—affect real-world FL performance metrics. Closing this knowledge gap through rigorous experimental analysis will empower the community to develop FL systems that are theoretically sound and empirically robust across application domains. 

Our study addresses this gap by using the \textsf{HD} to quantify differences in client data distributions, enabling a rigorous empirical analysis of non-IID effects across multiple dimensions, including label, feature, quantity, and spatiotemporal skews. Throughout the assessment of spatiotemporal skew, we capture the impact of dynamic data shifts over time and space, which are particularly relevant in real-world applications such as banking credit risk~\cite{zhang2024effects} and personalized healthcare~\cite{guo2025sthfl}. This approach ensures that our conclusions are robust and generalizable across diverse scenarios.

\paragraph{\textbf{Contribution.}}
The subsequent points encapsulate the contributions of our study:
\begin{enumerate}
  \item We benchmark five of the most employed state-of-the-art aggregation and client
  selection of FL algorithms to tackle \hl{non-IID data} distributions among
  clients under realistic, controlled, and quantifiable methods for
  synthetic data partitioning and all non-IID types (label, feature,
  quantity, and spatiotemporal skews). \hl{Previous empirical works have focused on label skew}~\cite{vahidian2023rethinking,wong2023empirical,mora2022federated}\hl{ or in combinations of label, feature, and quantity skew}~\cite{li2022federated} \hl{(see Section}\ref{sec:empirical_estudies}\hl{ for more details). Thus,} this is the first study
  empirically analyzing how the spatiotemporal skew affects the performance of FL models.
  \item We motivate using \textsf{HD} to quantify the differences among data distributions, standardizing the guidelines for systematic studies of \hl{non-IID data} in FL. This is the first work to demonstrate that the effect of the \hl{non-IID data} is not the same under all levels of heterogeneity.
  We use \textsf{HD} to quantify differences in
  distribution as it provides more granular
  information, and we leave as future work the
  exploration of other measures; see our section on design insights and opportunities.
  \item We provide a reference to researchers about which methods are robust to which kind of \hl{non-IID data} on highly benchmarked datasets.
  \item We give highlights and relevant recommendations for FL researchers based on quantifying the level of \hl{non-IID data}.
\end{enumerate}

To the best of our knowledge, this is the most comprehensive and
complete empirical study of \hl{non-IID data} and its effects on FL models.

\paragraph{\textbf{Positioning within Industrial Information Integration}.}
This study contributes to the Journal of Industrial Information Integration's focus on industrial \hl{non-IID data} and privacy-preserving analytics by thoroughly evaluating \hl{non-IID data} effects in FL. FL is increasingly adopted in industrial domains such as healthcare, intrusion detection in IoT, and digital twins for industrial IoT, where data is distributed across multiple silos and devices with inherent heterogeneity. Prior works in this journal have addressed related challenges, including emotion recognition based on electroencephalography (EEG) as a crucial research area in the Internet of Medical Things (IoMT)~\cite{jiang2025fuzzy}, federated ensemble model for intrusion detection in distributed IoT networks for enhancing cybersecurity~\cite{chahal2025design}, and adaptive optimization for FL enabled digital twins in industrial IoT~\cite{yang2024adaptive}. 

Our work advances these efforts by systematically quantifying the impact of diverse \hl{non-IID data} types on FL model performance using the HD metric. Furthermore, we benchmark state-of-the-art aggregation and client selection algorithms, offering practical guidance for deploying FL in industrial scenarios characterized by complex data distributions. This positioning situates our research within the ongoing scholarly conversation on industrial information integration and FL, underscoring its significance for robust, privacy-aware industrial analytics.

\paragraph{\textbf{Ethical Considerations in FL:}.}
\hl{FL inherently aligns with ethical principles related to data minimization and user privacy, as it allows individual clients to retain their raw data locally. However, despite these benefits, FL is not immune to ethical concerns. Potential privacy risks remain due to model inversion or gradient leakage attacks}~\cite{hatamizadeh2023gradient}\hl{, and there is a need for transparency and informed consent when deploying FL in real-world applications}~\cite{papadopoulos2021privacy}\hl{. Future work must integrate privacy-preserving mechanisms (e.g., differential privacy}~\cite{wei2020federated}\hl{, secure aggregation}~\cite{fereidooni2021safelearn}\hl{) and conduct rigorous audits to ensure ethical compliance in decentralized learning scenarios}~\cite{yuan2024digital}.

%% file: 02-section-related-work.tex
In this section, we present recent studies that empirically evaluate the behavior of non-IID data in FL. Additionally, we compare our work with relevant surveys on \hl{non-IID data} in FL.

\subsection{Empirical Studies} \label{sec:empirical_estudies}
Studies that analyze and benchmark the performance of methods to tackle the \hl{non-IID data} effect on the FL models under controlled and systematic scenarios are scarce. Nevertheless, in this section, we introduce those works that, to some extent, provide empirical analysis about the repercussions of \hl{non-IID data}.  

A study by Vahidian et al.~\cite{vahidian2023rethinking} challenges conventional thinking regarding \hl{non-IID data} in FL. They posit that dissimilar data among participants is not always detrimental and can be advantageous, and we found similar results. Their argument centers on two main points: firstly, that differences in labels (label skew) are not the sole determinant of \hl{non-IID data}, and secondly, that a more effective measure of heterogeneity is the angle between the data subspaces of participating clients. Complementary, \emph{we encompass a broader spectrum of \hl{non-IID data} types and include images and tabular data}.

Wong et al.~\cite{wong2023empirical} conduct extensive experiments on a large network of IoT and edge devices to present FL real-world characteristics, including learning performance and operation (computation and communication) costs. Moreover, they mainly concentrate on heterogeneous scenarios, the most challenging issue of FL. While they thoroughly analyze the impact of \hl{non-IID data}, the focus is primarily on image datasets, and they do not explore comparisons of aggregation algorithms to address highly heterogeneous scenarios. In contrast, \emph{our work expands on this by incorporating diverse datasets and benchmarking state-of-the-art methodologies to tackle \hl{non-IID data} in FL effectively}, offering a broader and more practical perspective.

In their study, Mora et al.~\cite{mora2022federated} examine existing solutions in the literature to mitigate the challenges posed by non-IID data. On the one hand, they emphasize the underlying rationale behind these alternative strategies and discuss their potential limitations. On the other hand, they identify the most promising approaches based on empirical results and critical defining characteristics, such as any assumptions made by each strategy. In addition, they focused on label skew and considered one dataset in their experiments. In our paper, \emph{we alternatively analyze broader datasets and data skewness types}, identifying limitations and potential approaches to overcome them.

Li et al.~\cite{li2022federated} conduct a comprehensive experimental evaluation of FL aggregation algorithms under non-IID data settings. They systematically analyze the strengths and limitations of state-of-the-art FL aggregation algorithms while introducing diverse data partitioning methods to simulate various non-IID scenarios. Their work highlights the challenges posed by non-IID data, such as accuracy degradation and training instability, and provides empirical insights into the performance of each algorithm across different settings. In our paper, \emph{we build upon their work by analyzing spatiotemporal skew and introducing metrics to quantify its effects}, offering a broader perspective on \hl{non-IID data} and its impact on model performance.

\subsection{Surveys}

Some surveys that explore the effects of \hl{non-IID data} in the model performance have been proposed, and they are depicted in Table~\ref{tab:surveys_comparison}. Earlier works, such as those from 2024, focus on label skew, providing valuable insights into this dimension. Resources from 2022 partially mention spatiotemporal skew, contributing to a deeper understanding of these aspects. The 2021 study offers an essential foundation by addressing label skew, paving the way for more comprehensive analyses in subsequent years.

\begin{table}[H]
  
  \resizebox{\columnwidth}{!}{%
    \begin{tabular}{||c|c|c|c|c|c|c|c|c|c||}
      \hline
      \textbf{\makecell{Resource}} & \textbf{\makecell{Publication \\ Year}} & \textbf{\makecell{Label \\ Skew}} & \textbf{\makecell{Feature  \\ Skew}} & \textbf{\makecell{Quantity \\ Skew}} & \textbf{\makecell{Spatiotemporal \\ Skew}} & \textbf{\makecell{\hl{Non-IID data}\\ Quantification}} & \textbf{\makecell{Empirical\\Highlights}} \\ \hline\hline

      \textbf{\cite{lu2024federated}} & 2024 & \textcolor{teal}{\faCheckCircle} & \textcolor{purple}{\faTimesCircle} & \textcolor{purple}{\faTimesCircle} & \textcolor{purple}{\faTimesCircle} & \textcolor{purple}{\faTimesCircle} & \textcolor{purple}{\faTimesCircle} \\ \hline

      \textbf{\cite{pei2024review}} & 2024 & \textcolor{teal}{\faCheckCircle} & \textcolor{teal}{\faCheckCircle} & \textcolor{teal}{\faCheckCircle} & \textcolor{purple}{\faTimesCircle} & \textcolor{orange}{\faMinusCircle} & \textcolor{purple}{\faTimesCircle} \\ \hline

      \textbf{\cite{ma2022state}} & 2022 & \textcolor{teal}{\faCheckCircle} & \textcolor{teal}{\faCheckCircle} & \textcolor{teal}{\faCheckCircle} & \textcolor{purple}{\faTimesCircle} & \textcolor{purple}{\faTimesCircle} & \textcolor{purple}{\faTimesCircle} \\ \hline

      \textbf{\cite{criado2022non}} & 2022 & \textcolor{orange}{\faMinusCircle} & \textcolor{purple}{\faTimesCircle} & \textcolor{purple}{\faTimesCircle} & \textcolor{teal}{\faCheckCircle} & \textcolor{purple}{\faTimesCircle} & \textcolor{purple}{\faTimesCircle} \\ \hline
      
      \textbf{\cite{types_noniid2}} & 2021 & \textcolor{teal}{\faCheckCircle} & \textcolor{teal}{\faCheckCircle} & \textcolor{teal}{\faCheckCircle} & \textcolor{orange}{\faMinusCircle} & \textcolor{purple}{\faTimesCircle} & \textcolor{purple}{\faTimesCircle} \\ \hline
      
      \textbf{Ours} & 2025 & \textcolor{teal}{\faCheckCircle} & \textcolor{teal}{\faCheckCircle} & \textcolor{teal}{\faCheckCircle} & \textcolor{teal}{\faCheckCircle} & \textcolor{teal}{\faCheckCircle} & \textcolor{teal}{\faCheckCircle} \\      
      \hline
    \end{tabular}%
  }
    \caption{Comparison against surveys (resources) for \hl{non-IID data} in FL (\textcolor{teal}{\faCheckCircle}: Included, \textcolor{orange}{\faMinusCircle}: Partially included, \textcolor{purple}{\faTimesCircle}: Not included)}
    \label{tab:surveys_comparison}
\end{table}

Compared to the previous surveys summarized in Table~\ref{tab:surveys_comparison}, our work builds upon and extends their contributions by offering a more comprehensive approach. \emph{We include empirical evaluations of the effects of non-IID data, provide quantification of the \hl{non-IID data} level, and conduct extensive experiments to assess the impact of spatiotemporal skew thoroughly}. These additions enhance our study's depth and practical relevance, complementing earlier works.

%% file: 03-section-background.tex
This section provides an overview of FL, its fundamental training process, and the challenges posed by \hl{non-IID data}. We categorize different types of data skew that impact FL performance and introduce methods for quantifying \hl{the level of non-IID data}. Additionally, we review state-of-the-art aggregation and client selection strategies designed to address these challenges, such as FedAvg~\cite{mcmahan2017communication}, FedProx~\cite{li2020federated}, Random size-proportional selection (Rand)~\cite{cho2022towards}, Power-Of-Choice (POC)~\cite{cho2022towards}, and Model Contrastive Learning (MOON)~\cite{li2021model}.

\subsection{Basics of FL}

FL~\cite{mcmahan2017communication} suits siloed data (a.k.a. clients, local nodes, parties, participants) where multiple organizations or institutions hold their datasets. This decentralized approach enables collaborative model training without sharing the raw data, contributing to data privacy and ownership for each participating organization~\cite{elayan2021deep}. 

\begin{figure}[ht]
\centering
\includegraphics[width=0.99\columnwidth]{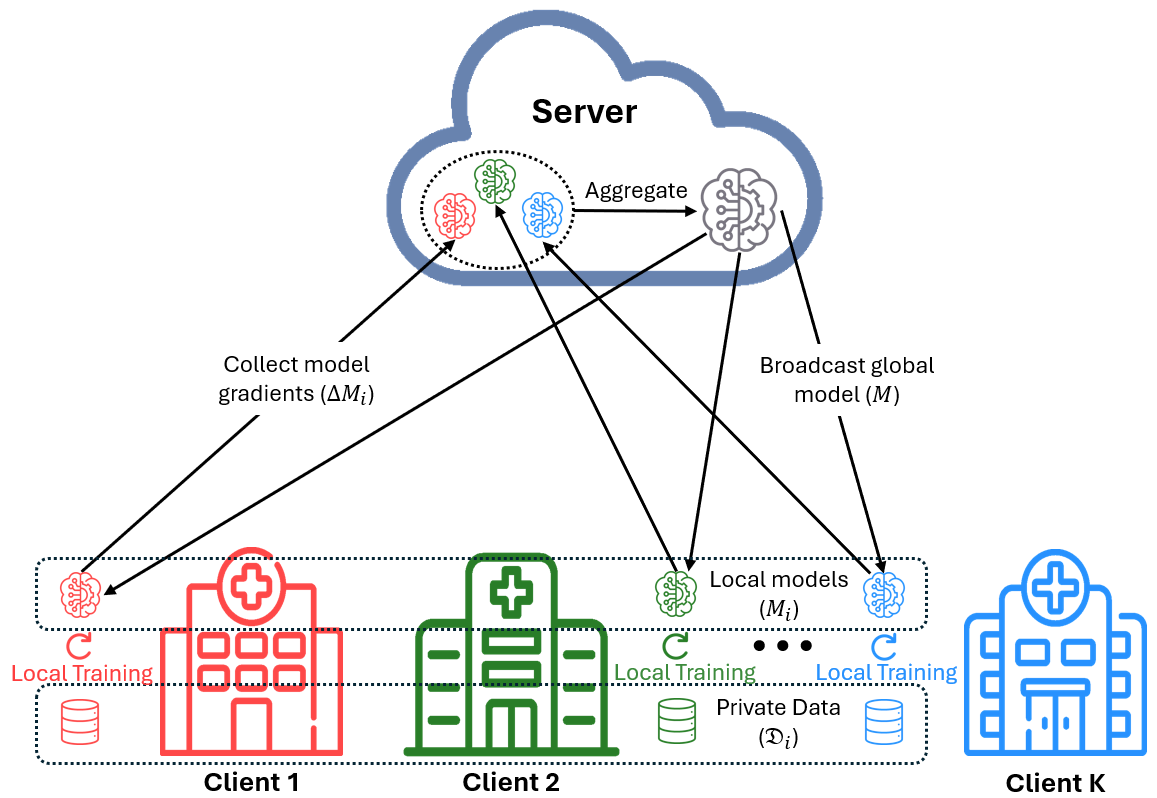}
\caption{FL training process overview}
\label{fig:fl_overview}
\end{figure}

Figure~\ref{fig:fl_overview} presents an overview of the cross-silo FL training process, where participating clients train local models
($M_i$) on their datasets ($\mathcal{D}_i$), all based on a
pre-distributed global model
($M$)~\cite{sakib2021asynchronous,zhang2020federated}. Instead of
sharing raw data, clients exchange model updates ($M_i$) without
sensitive information, aggregating centrally to enhance the global
model. By utilizing a central server for coordination, clients transmit
their updates, aggregated to improve the global model. This iterative
process allows collaboration without compromising data privacy, as
each client receives the updated global model without sharing raw data, ensuring data privacy while facilitating collaboration.

\subsection{Data skew types.}

A \emph{centralized dataset}\footnote{Notice that this definition of a centralized dataset includes tabular data, images, medical data, and graph data, and any dataset expressible as a collection of arrays.} $\mathcal{D}=\{(\mathbf{x}_1,y_1),\dots,(\mathbf{x}_n,y_n)\}$, is a collection of $n$ tuples where $\mathbf{x}_i=[(x_i)_1,\dots,(x_i)_m]$ is the feature representation of the $i$th element (sample) in the dataset, and $y_i\in\{1,\dots,\ell\}$ is the (true) label of the $i$th element.

In the FL setting, the dataset $\mathcal{D}$ is distributed over $K$
clients. We let $\mathcal{D}_i$ be the set of elements of the $i$th
client. That is:
\[\mathcal{D}=\cup_{i=1}^K\mathcal{D}_i\qquad\text{and for $i\ne j$:}\qquad
\mathcal{D}_i\cap\mathcal{D}_j=\emptyset.\]
Defining the type of \hl{non-IID data} in FL is relevant since it can
drastically influence the performance of the models. We follow the
settings of previous work~\cite{types_noniid1,types_noniid2}.
For a supervised learning task on
client $i$ (local node $i$), we assume that each data sample $(x,
y)\in\mathcal{D}_i$, where $x$
is the input attributes or features, and $y$ is the label, following a
local distribution $P_i(\mathbf{x}, y)$.
Let us define:

\begin{equation} \label{eq:def_lab_quant}
    P_i^Y(y) = \sum_{\substack{(\mathbf{x},z)\in\mathcal{D}_i\!\!\\z=y}} P_i(\mathbf{x},z)
    \quad \text{and} \quad
    P_i^{X_\ell}(x) = \sum_{\substack{(\mathbf{x},y)\in\mathcal{D}_i\!\!\\x_\ell=x}} P_i(\mathbf{x},y)
\end{equation}
\noindent
with $P_i^Y(y)$, the $i$th client labels' distribution and
$P_i^{X_\ell}(x)$ the distribution over the $\ell$th input feature of
the $i$th client. Then, the classification for non-IID data (i.e., data
skew types) is as follows:

\begin{itemize}
  \item Regarding the concept of \textit{identically distributed}:
\begin{enumerate}
  \item \textbf{Label skew}: Means that the label distribution $P_i^Y(y)$ of different clients is different.
  \item \textbf{Feature skew}: Occurs when the distribution of the
  features $P_i^{X_\ell}(x)$ varies from client to client.
  \item \textbf{Quantity skew}: Refers to the significant difference
  in the number of examples of different client data $P_i(\mathbf{x}, y)$.
\end{enumerate}

\item Regarding the concept of \textit{independent}:
    \begin{enumerate}\addtocounter{enumi}{3}
\item \textbf{Spatiotemporal skew}: Also known as spatial-temporal skew under federated continual learning (FCL)~\cite{yu2024overcoming,yang2024federated,yu2024addressing}. It refers to the inner correlation of
data in the time (or space) domain. In other words, the distribution
$P_i(\mathbf{x},y)$ is not stationary but depends on time or space. 

\end{enumerate}
\end{itemize}

\subsection{Quantifying the Degree of non-IID data.}

Regarding selecting valuable scenarios to demonstrate the effects of \hl{non-IID data} in FL, the current literature often relies on ad-hoc partitions~\cite{dirichlet_method,aggregators_FL,percent_non_iid_method}. Therefore, in this work, we use a \textit{metric that systematically evaluates the level of non-IID data to select scenarios for measuring the effect of \hl{non-IID data}}. We opted for the Hellinger Distance (\textsf{HD}), a metric widely used to gauge the separation between two probability distributions calculated as in Equation~\ref{eq:hd_equation}~\cite{hd_properties}. 

\begin{equation} \label{eq:hd_equation}
    \resizebox{0.47\textwidth}{!}{$
    \textsf{HD}(P_1^Y(y),P_2^Y(y)) =
    \frac{1}{\sqrt{2}}\sqrt{\sum_{y\in Y}\left(\sqrt{P_1^Y(y)}-\sqrt{P_2^Y(y)}\right)^2}$}
\end{equation}

HD provides a fine-grained and sensitive measurement of distributional differences, reaching values close to 1 under extreme non-IID conditions—unlike JSD, which often saturates and fails to reflect high levels of skew. Additionally, HD is highly adaptable across different types of \hl{non-IID data}. In contrast to Earth Mover’s Distance (EMD)~\cite{jimenez2024fedartml}, whose values depend heavily on the choice and scale of the ground distance, HD offers normalized and consistent comparisons across tasks and datasets, making it particularly well-suited for FL scenarios.

\subsection{Aggregation and Client Selection Algorithms}

In an FL process, the server aggregates the weights obtained from each
client and communicate them back to each participant. 

In this section, we explain the five state-of-the-art aggregation
and client-selection algorithms assessed in our experiments. 

\emph{FedAvg:} It is a fundamental algorithm in FL~\cite{mcmahan2017communication} designed to train ML models across a
network of decentralized devices while preserving data privacy. In
FedAvg, each client computes model updates and sends them to a central
server using local data. The server averages these updates to calculate
a global model update and then sends it back to the clients. This 
process iterates until it converges (the model's performance gets
stable). \hl{FedAvg suffers from three key limitations: (1) degraded performance under non-IID data distributions across clients, (2) inefficient communication rounds caused by straggling devices or disproportionate local dataset sizes, and (3) a uniform aggregation approach that fails to account for variability in client data quality or device reliability, potentially biasing the global model.}

\emph{FedProx:} It is a framework designed to address \hl{non-IID data} in
FL~\cite{li2020federated}, offering a generalized and reparametrized
version of FedAvg. This approach incorporates a regularization term
($\mu$) to minimize the difference between local and global weights.
Finally, the framework aggregates local model updates from all devices
to obtain an updated global model. Using this proximal term, FedProx
aims to improve convergence and performance in heterogeneous federated
learning environments. \hl{FedProx ensures convergence even with non-IID data while requiring only minor adjustments to implementation. However, it has notable drawbacks: (1) its effectiveness depends significantly on careful hyperparameter selection, and incorrect settings may slow convergence or raise communication overhead; (2) in real-world applications, its advantages weaken under extreme levels of non-IID data, particularly when client datasets contain completely distinct classes.}~\cite{li2022federated}. 

\emph{Rand:} Rand is a baseline client selection strategy introduced to handle non-IID data that is not biased toward clients with higher local losses. Most current analysis frameworks consider a scheme that selects the training set of clients $S(t)$ by sampling $m$ clients randomly (with replacement) such that client $k$ gets selected with probability $p_k$, the fraction of data at that client~\cite{cho2020client}. \hl{Rand provides unbiased client selection but suffers from key limitations: (1) it fails to prioritize clients with informative updates, hindering convergence speed; and (2) in highly non-IID settings, it may underrepresent rare data distributions, reducing model generalization.}

\emph{POC:} The POC algorithm performs well under a non-IID distribution. It is inspired by the power of
$d$ choices load balancing strategy, which queueing systems commonly
use~\cite{cho2022towards,cho2020client}. The central server
first samples a candidate set of $d$ clients, where $d$ is between $m$
(the number of clients to be selected) and $K$. These candidates are
chosen based on their data fraction ($p_k$). The server then sends the
current global model to these candidates, who compute and return their
local losses. Finally, the server selects $m$ clients with the highest
losses to participate in the next training round. This approach aims to
balance the workload and prioritize clients with more informative
updates, improving the efficiency of the FL process. \hl{While this approach enhances training efficiency and manages non-IID data effectively, it also has limitations: (1) Selection bias can marginalize underrepresented clients, thereby weakening the model’s generalization ability. (2) The method introduces higher complexity and greater communication costs in the client selection process.}

\emph{MOON:} It is a simple and effective FL framework designed to tackle \hl{non-IID data}. It uses the similarity between model representations to correct the local training of individual parties (i.e., conducting contrastive learning at the model level). The network proposed in MOON has three components: a base encoder, a projection head, and an output layer. The base encoder extracts representation vectors from inputs. Le et al.~\cite{li2021model} introduce an additional
projection head to map the representation to a space with a fixed
dimension. Last, the output layer produces predicted values for each
class. For ease of presentation, with model weight $w$, they use
$F_w(\cdot)$ to denote the whole network and $R_w(\cdot)$ to denote the
network before the output layer (i.e., $R_w(X_\ell)$ is the mapped
representation vector of input $X_\ell$). \hl{Like the previous methods, this approach is effective but has drawbacks: (1) higher computational costs due to generating and comparing augmented data views, and (2) restricted use for non-visual data (e.g., text or time-series), where creating meaningful augmentations is difficult because of text context-dependence and time-series structural limitations.}~\cite{chen2023pfl}

\subsection{Models used in FL}

\hl{In FL, the choice of model architecture plays a critical role in determining both performance and communication efficiency across distributed clients. Depending on the nature of the data and the target task, different types of models may be employed to balance expressiveness, computational cost, and generalizability}~\cite{wu2022communication}\hl{. Below, we outline several common model types used in FL, highlighting their core characteristics and suitability for decentralized training environments.}

\begin{itemize}
    \item \hl{\emph{Deep Neural Networks (DNNs)}: DNNs are feedforward networks with multiple hidden layers, capable of learning complex patterns through hierarchical feature extraction}~\cite{alhalabi2023fednets}\hl{. They serve as foundational models in FL due to their flexibility.}
    \item \hl{\emph{Convolutional Neural Networks (CNNs)}: CNNs specialize in processing grid-like data (e.g., images) using convolutional layers for local feature detection, pooling for dimensionality reduction, and fully connected layers for classification}~\cite{he2020group}\hl{. Their parameter-sharing property makes them efficient for FL tasks.}
    \item \hl{\emph{Transfer Learning Models}: Pre-trained architectures like ResNet9}~\cite{resnet9}\hl{, EfficientNetB0}~\cite{tan2019efficientnet}\hl{, and MobileNetV2}~\cite{sandler2018mobilenetv2}\hl{, leverage transfer learning by adapting learned features from large datasets (e.g., ImageNet) to new tasks with limited data}~\cite{wang2022progfed}\hl{. In FL, such models reduce communication overhead and improve convergence by starting from robust initial weights.}
\end{itemize}

%% file: 04-section-experiments.tex
\paragraph{Datasets.}
This work considers eight widely employed real datasets to train the centralized learning (CL) and FL models. Four of these, i.e., CIFAR10~\cite{cifar10},
FMNIST~\cite{xiao2017fashion}, Physionet 2020~\cite{gutierrez2022application}, and Covtype~\cite{misc_covertype_31} serve to simulate label, feature, and quantity skew. To further examine label skew in scenarios with a considerably larger number of classes, we also included CIFAR100~\cite{krizhevsky2009learning}.
The remaining three datasets, i.e., 5G Network Traffic flows~\cite{choi2023ml},  MHEALTH~\cite{mhealth_319}, and Snapshot Serengeti~\cite{swanson2015snapshot}, are used to simulate spatiotemporal skew. Table~\ref{tab:characteristics_datasets} provides an overview of the main characteristics of each dataset.

\begin{table}[htbp]
    \centering
    \caption{Characteristics of the datasets} 
    \resizebox{\columnwidth}{!}{
    \begin{tabular}{||c|c|c|c|c|c|c||}
    \hline
    Dataset & Type & \makecell{\#training \\examples} & \makecell{\#test \\examples} & \#features & \#classes & \makecell{Classes \\ distribution} \\
    \hline\hline
    CIFAR10 & Images & 50,000  & 10,000 & 3,072 & 10 & Balanced \\
    \hline
    FMNIST & Images & 60,000  & 10,000 & 784 & 10 & Balanced \\
    \hline
    CIFAR100 & Images & 50,000  & 10,000 & 3,072 & 100 & Balanced \\
    \hline
    Physionet & Tabular & 39,895  & 2,095 & 120 & 27 & Balanced \\
    \hline
    Covtype & Tabular & 522,910  & 58,102 & 54 & 7 & Unbalanced \\
    \hline
    Serengeti & Tabular & 257,927 & 28,659 & 64 & 13 & Unbalanced \\
    \hline
    5G NTF & Tabular & 74,838 &  13,207 & 7 & 12 & Unbalanced \\
    \hline
    MHEALTH & Tabular & 851,021 &  364,724 & 14 & 13 & Unbalanced \\
    \hline
    \end{tabular}}
    \label{tab:characteristics_datasets}
\end{table}

\paragraph{Models.}

We adopt a well-studied CNN broadly applied in computer vision~\cite{CNN_inspiration} for the CIFAR10 and FMNIST datasets. It includes one input layer and three convolutional blocks, where the first two blocks each have a convolutional layer followed by a max pooling layer, and the final block contains a convolutional layer and a flattened layer. The initial convolutional layer has 32 filters, whereas the subsequent two layers each have 64 filters with a $3\times 3$ filter size and ReLu as the activation function. In the dense section of the network, there is one dense layer with 64 neurons using ReLU as the activation function. We utilized a ResNet9 for the CIFAR100 dataset. Additionally, we employed in our tests the transfer learning models EfficientNetB0 and MobileNetV2 since they produce higher classification power results for the datasets studied.

For the tabular datasets, we use a DNN, selected
because it is widely employed in classification tasks with tabular
data~\cite{saeed2024ecg}. It comprises one input layer, three hidden
layers, and one output layer~\cite{inspiration_DNN}. The input layer
uses as many units as the number of features in the training set. The three layers contain 500 hidden units each, and the last layer is formed by considering the neurons equal to the number of classes to predict. Additionally, the hidden layers used the ReLu activation function, and the output layer used a SoftMax function. We use Adam as our optimizer with a learning rate of 0.001 for $K=30$ clients and a batch size of 64. In our simulations, models were trained for 40 communication rounds and 10 local epochs, except for those using the MOON aggregation algorithm and those involving the CIFAR100 dataset, which were trained for 100 communication rounds to ensure convergence in those settings. To ensure reproducibility and statistical validity, we executed all experiments across the datasets using five and ten distinct data partitions generated from fixed random seeds. 

\subsection{Hyperparameters Tuning}
For a fair comparison, we base our hyperparameter grids on the best-performing hyperparameters presented in the original papers as follows: 

\begin{itemize}
  \item \textbf{FedAvg}: We do not set any specific tuning process for this algorithm~\cite{mcmahan2017communication}. 
  \item \textbf{Rand}: The fraction of clients considered in each
  communication round gets fine-tuned from
  $\{0.3, 0.5, 0.7\}$~\cite{cho2020client}.
  \item \textbf{FedProx}: The $\mu$  parameter gets fine-tuned from \{0,
  0.001, 0.01, 0.1, 1, 10, 100\}~\cite{li2020federated}.
  \item \textbf{POC}: The parameter $C$ is equal to 0.5. The parameter
  $d$ gets fine-tuned from
  $\{15, 18, 19, 21\}$~\cite{cho2020client}.
  \item \textbf{MOON}: The $\mu$ is tuned from the grid of
  $\{0.1, 1, 5, 10\}$, and we find the best $\mu$ of 0.1, and we set the
  value of $temperature$ to 0.5~\cite{li2021model}.
\end{itemize}

\subsection{Hardware Specification}
We used an Ubuntu 22.04.4 LTS machine with 200 GB of disk, Intel(R)
Xeon(R) Platinum 8259CL CPU @ 2.50GHz processor, 16 processors, 125 GB
of RAM, and Python 3.10.12 to run the experiments. The FL models were trained using the Flower~\cite{beutel2020flower} platform.

\subsection{Performance Metrics}
This subsection describes the performance and convergence metrics considered in our experiments and a justification for their use. \\

\emph{Accuracy~\cite{vahidian2023rethinking,wong2023empirical,mora2022federated}.} It refers to the proportion of correctly classified instances compared to the total data size. Higher values of accuracy indicate a better model performance. It can be calculated as follows:
\begin{equation}
    \hspace*{\fill}
    Acc = \frac{\sum_{k=1}^{K} C_k}{\sum_{k=1}^{K} n_k}
    \hspace*{\fill}
\end{equation}
\noindent
where $C_k$ indicates the number of correctly classified samples on client $k$ and $n_k$ is the number of data samples on client $k$. We performed five and ten independent trials using different random seeds for the experiments. To ensure robust and reliable results, we report the mean accuracy and the standard deviation across these trials, providing a comprehensive view of the model's performance variability.

\emph{\hl{Curvature}~\cite{docarmo2016differential, gray2006modern}.} 
\hl{We incorporate curvature as a metric to identify points along the accuracy curve with respect to \textsf{HD} where model performance changes considerably. Given a parametric curve  $\alpha(t) = (x(t), y(t))$, where $x(t)$ denotes the level of non-IID data quantified by \textsf{HD} and $y(t)$ the corresponding model accuracy, the curvature $\kappa(t)$ at that point is defined as:}
\begin{equation}
    \hspace*{\fill}
    \kappa(t) = \frac{x'(t)y''(t) - x''(t)y'(t)}{\left(x'(t)^2 + y'(t)^2\right)^{3/2}}
    \hspace*{\fill}
\end{equation}
\hl{where $x'(\cdot)$ denotes the first-order derivative and $x''(\cdot)$ is the second-order derivative. 

\emph{Number of times detected as critical point (\#Detected as critical point).} To identify critical points related to the effects of non-IID data, we count how many times each \textsf{HD} value is detected as critical based on curvature. Specifically, points where the curvature satisfies $\kappa \ge 1$ are considered indicators of sharp performance degradation. This count-based metric highlights \textsf{HD} values where degradation occurs consistently across the different models built by varying the label skew of the clients' data distributions, reflecting the robustness and consistency of a critical point of change across different settings.

\emph{Average curvature.} This metric captures the overall sharpness of performance change at each \textsf{HD} value by averaging curvature values across different models built under the same non-IID data. By averaging out the curvature values across models, it offers a more robust and reliable estimate of how critical each point truly is. A higher average curvature indicates a sharper and more consistent performance drop, pointing to greater sensitivity or instability at that level of non-IID data. This metric aims to identify the points where performance degradation not only present but also most pronounced, helping us more effectively evaluate and compare the flagged \textsf{HD} values (0.5 and 0.75) as critical points.}

\emph{Number of times that performed the best~\cite{li2022federated}.} This metric quantifies how frequently an aggregation algorithm performs better than other approaches across multiple experimental trials. A higher count indicates greater consistency and robustness. This metric helps identify which methods consistently excel across different data partitions, which is particularly valuable in FL, where \hl{non-IID data} distributions can lead to high-performance variance between trials.

\emph{Rounds-to-accuracy (RTA)~\cite{wu2021fast}.} This metric measures the minimal number of communication rounds needed for the global model to achieve at least 90\% of the maximum accuracy among the aggregation algorithms. It reflects the efficiency of the FL process since lower values indicate a faster model's convergence.

%% file: 05-section-label-results.tex
This section examines how label skew in the client data affects the models' performance. Consider that the Covtype is an unbalanced dataset regarding the labels, and CIFAR10, FMNIST, CIFAR100, and Physionet are balanced datasets. The accuracy of the models on each CL is our baseline for comparing the accuracy of the models generated in FL. 

\subsection{Synthetic Partitioning Method}
Using the FedArtML tool~\cite{jimenez2024fedartml}, we employed the Dirichlet distribution (DD) to partition data among clients based on label distribution. The DD generates random numbers summing to one, controlled by the parameter $\alpha$. Higher $\alpha$ values (e.g., 1000) create similar local distributions, while lower values increase the chance of clients having examples from a single, randomly chosen class~\cite{dirichlet_method}. \hl{The selected values \{1000, 6, 1, 0.3, 0.03\} allow us to examine the impact of varying degrees of non-IID data on FL performance.} Notice that the DD is the multivariate generalization of the Beta distribution, \hl{and the Beta distribution is itself a generalization of the Uniform distribution.}. Therefore, the partition of the datasets using DD is a skewed split of the data distribution~\cite{lin2016dirichlet}. 

\begin{figure}[ht]
\centering
\includegraphics[width=0.99\columnwidth]{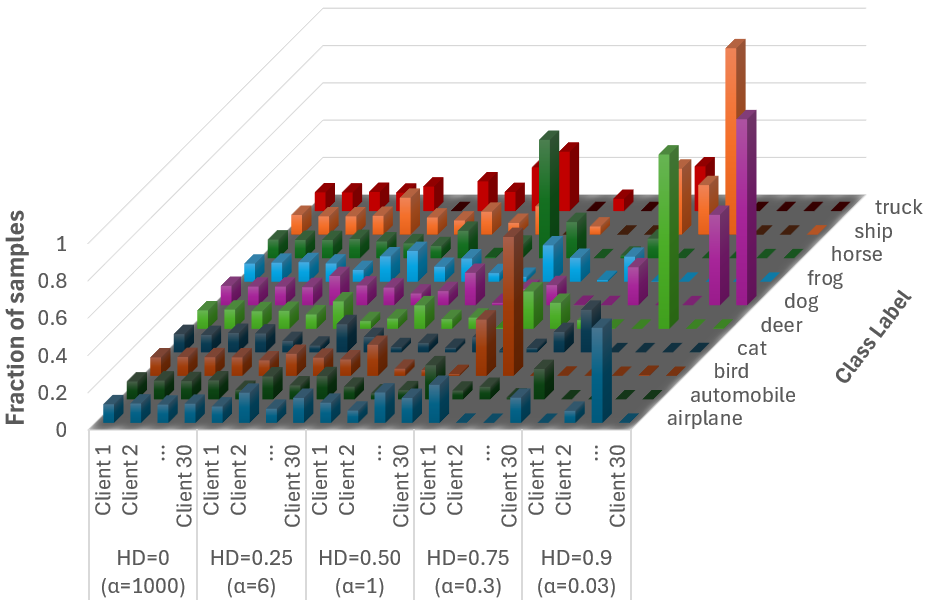}
\caption{Distribution of CIFAR10 among 30 clients for different levels of \hl{non-IID data}. The x-axis shows distinct $\alpha$ values used to partition the data and the resulting \textsf{HD} for clients from 1 to 30. The y-axis shows the participation of each class depicted on the z-axis.}
\label{fig:example_label_skew_dirichlet_30cli}
\end{figure}

\hl{We quantify the degree of non-IID data across clients using the \textsf{HD}. Specifically, we partition the data using the DD's $\alpha$ values $\{1000, 6, 1, 0.3, 0.03\}$ to achieve distinct \textsf{HD} levels $\{0.0, 0.25, 0.5, 0.75, 0.9\}$ to cover a representative spectrum of non-IID data ranging from fully IID to highly non-IID partitions. For instance:}

\begin{itemize}
    \item \hl{A DD concentration parameter $\alpha=1000$ yields IID data ($\textsf{HD} \approx 0.0$), as labels are uniformly distributed.}
    \item \hl{Conversely, $\alpha=0.03$ produces highly skewed partitions ($\textsf{HD} \approx 0.9$).}
\end{itemize}

\hl{Thus, each $\alpha$ value maps to a unique \textsf{HD} based on the label distribution, enabling controlled experimentation across non-IID scenarios.} Figure~\ref{fig:example_label_skew_dirichlet_30cli} exemplifies the
partition distribution for label skew using thirty clients. 
All ten classes get evenly distributed among every client in the IID
scenario ($\alpha=1000, \textsf{HD}=0.0$). As we increase the $\alpha$
parameter in the DD, the distribution of classes among clients becomes
more diverse. In the extreme case of $\alpha = 0.03, \textsf{HD}=0.9$, certain
classes are absent in some clients.

\subsection{Classification Power}
In this subsection, we focus on the findings from the simulations to compare different aggregation algorithms and datasets regarding their classification power (a.k.a. accuracy). 

\begin{table*}[htbp]
    \centering
    \caption{Mean and standard deviation Accuracy for each dataset for CL, FedAvg, Rand, FedProx, Power-Of-Choice, and MOON, considering different levels of \hl{non-IID data} as measured by \textsf{HD} for $K = 30$. Each model has undergone ten different trials (random seeds). }
    \resizebox{\textwidth}{!}{
    \begin{tabular}{||c|c|c|c|c|c|c|c|c|c||}
    \hline
    Category& Dataset & \textsf{HD} & CL & FedAvg & Rand & FedProx & POC & MOON\\
    \hline\hline

    \multirow[c]{25}{*}{ \rotatebox[origin=c]{90}{Label distribution skew}} & \multirow[c]{5}{*}{CIFAR10}& 0 & \multirow[c]{5}{*}{70.50\% $\pm$ 0.60\%} & 66.12\% $\pm$ 0.73\% & 66.16\% $\pm$ 0.74\%  & \textbf{66.35\% $\pm$ 0.72\%} & 66.26\% $\pm$ 0.70\% & 64.45\% $\pm$ 1.05\%  \\ \cline{3-3}\cline{5-9}
    
    && 0.25 && \textbf{65.91\% $\pm$ 0.49\%} & 65.49\% $\pm$ 0.52\%  & 65.86\% $\pm$ 0.60\% & 65.61\% $\pm$ 0.67\% & 63.40\% $\pm$ 0.74\% \\ \cline{3-3}\cline{5-9}
    
    && 0.5 && 63.41\% $\pm$ 0.95\% & 62.93\% $\pm$ 1.55\%  & \textbf{63.56\% $\pm$ 0.83\%} & 62.86\% $\pm$ 1.71\% & 60.95\% $\pm$ 1.33\%  \\ \cline{3-3}\cline{5-9}
    
    && 0.75 && \textbf{58.85\% $\pm$ 1.06\%} & 56.80\% $\pm$ 2.24\%  & 58.80\% $\pm$ 1.04\% & 55.84\% $\pm$ 1.48\% & 55.27\% $\pm$ 0.51\% \\ \cline{3-3}\cline{5-9}
    
    && 0.9 && 43.22\% $\pm$ 2.24\% & 40.95\% $\pm$ 2.43\%  & \textbf{44.33\% $\pm$ 2.83\%}  & 39.04\% $\pm$ 2.99\% & 38.84\% $\pm$ 2.31\% \\ \cline{2-9}

    & \multirow[c]{5}{*}{FMNIST}& 0 & \multirow[c]{5}{*}{90.90\% $\pm$ 0.20\%} & 90.68\% $\pm$ 0.18\% & 90.63\% $\pm$ 0.18\% & \textbf{90.69\% $\pm$ 0.15\%} &  90.62\% $\pm$ 0.17\% & 88.70\% $\pm$ 0.27\%  \\ \cline{3-3}\cline{5-9}
    
    && 0.25 && 90.44\% $\pm$ 0.14\% & \textbf{90.52\% $\pm$ 0.17\%} & 90.51\% $\pm$ 0.17\% &  90.51\% $\pm$ 0.18\% & 88.17\% $\pm$ 0.22\% \\ \cline{3-3}\cline{5-9}

    && 0.5 && 89.92\% $\pm$ 0.11\% & 89.84\% $\pm$ 0.31\% & \textbf{89.96\% $\pm$ 0.20\%} &  89.74\% $\pm$ 0.35\% & 87.37\% $\pm$ 0.22\% \\ \cline{3-3}\cline{5-9}
    
    && 0.75 && 88.15\% $\pm$ 0.54\% & 87.51\% $\pm$ 0.81\% & \textbf{88.17\% $\pm$ 0.47\%} & 87.23\% $\pm$ 0.78\%  & 84.70\% $\pm$ 0.84\% \\ \cline{3-3}\cline{5-9}
    
    && 0.9 && 80.37\% $\pm$ 3.78\% & 79.08\% $\pm$ 4.67\% & \textbf{81.10\% $\pm$ 2.21\%} & 77.83\% $\pm$ 3.28\%  & 70.79\% $\pm$ 5.73\% \\ \cline{2-9}

    & \multirow[c]{5}{*}{CIFAR100}& 0 & \multirow[c]{5}{*}{67.47\% $\pm$ 0.46\%} & 62.88\% $\pm$ 0.28\% & 62.72\% $\pm$ 0.35\% & \textbf{63.05\% $\pm$ 0.12\%} &  62.80\% $\pm$ 0.38\% & 56.51\% $\pm$ 0.30\%  \\ \cline{3-3}\cline{5-9}
    
    && 0.25 && \textbf{62.66\% $\pm$ 0.35\%} & 62.45\% $\pm$ 0.25\% & 62.55\% $\pm$ 0.32\% &  62.30\% $\pm$ 0.21\% & 56.32\% $\pm$ 0.49\% \\ \cline{3-3}\cline{5-9}

    && 0.5 && 61.72\% $\pm$ 0.60\% & 61.49\% $\pm$ 0.39\% & \textbf{61.74\% $\pm$ 0.26\%} &  61.23\% $\pm$ 0.17\% & 56.47\% $\pm$ 0.19\% \\ \cline{3-3}\cline{5-9}
    
    && 0.75 && 59.47\% $\pm$ 0.37\% & 58.46\% $\pm$ 0.73\% & \textbf{59.72\% $\pm$ 0.51\%} & 59.10\% $\pm$ 0.25\% & 56.24\% $\pm$ 0.42\% \\ \cline{3-3}\cline{5-9}
    
    && 0.9 && 54.38\% $\pm$ 0.69\% & 52.87\% $\pm$ 1.17\% & \textbf{54.80\% $\pm$ 0.63\%} & 52.85\% $\pm$ 0.99\% & 51.45\% $\pm$ 1.18\% \\ \cline{2-9}
    
    & \multirow[c]{5}{*}{Physionet}& 0 & \multirow[c]{5}{*}{63.74\% $\pm$ 1.24\%}& 57.97\% $\pm$ 0.49\% & 57.48\% $\pm$ 0.40\%  & 58.16\% $\pm$ 0.62\% & 57.86\% $\pm$ 0.76\% & \textbf{61.80\% $\pm$ 0.62\%} \\ \cline{3-3}\cline{5-9}

    && 0.25 && 57.65\% $\pm$ 0.47\% & 57.42\% $\pm$ 0.54\%  & 57.79\% $\pm$ 0.55\% & 57.48\% $\pm$ 0.53\% & \textbf{60.94\% $\pm$ 0.60\%} \\ \cline{3-3} \cline{5-9}

    && 0.5 && 55.69\% $\pm$ 0.93\% & 55.26\% $\pm$ 1.05\%  & 56.29\% $\pm$ 1.00\% & 55.24\% $\pm$ 1.48\% & \textbf{58.76\% $\pm$ 0.71\%} \\ \cline{3-3} \cline{5-9}
    
    && 0.75 && 50.88\% $\pm$ 1.18\% & 50.19\% $\pm$ 2.07\%  & 51.47\% $\pm$ 1.20\% & 49.51\% $\pm$ 2.30\% & \textbf{53.51\% $\pm$ 1.37\%}  \\ \cline{3-3}\cline{5-9}
    
    && 0.9 && 41.35\% $\pm$ 2.70\% & 39.68\% $\pm$ 3.01\%  & 41.95\% $\pm$ 2.07\% & 38.81\% $\pm$ 3.68\% & \textbf{42.49\% $\pm$ 3.00\%}  \\  
    \cline{2-9}
    
    & \multirow[c]{5}{*}{Covtype}& 0 & \multirow[c]{5}{*}{95.60\% $\pm$ 0.10\%} & 94.95\% $\pm$ 0.06\% & 94.89\% $\pm$ 0.08\% & 94.96\% $\pm$ 0.09\% & 94.84\% $\pm$ 0.10\%  & \textbf{95.64\% $\pm$ 0.06\%}  \\ \cline{3-3}\cline{5-9}
    
    && 0.25 && 92.05\% $\pm$ 1.14\% & 91.63\% $\pm$ 1.52\% & 92.10\% $\pm$ 1.28\% & \textbf{93.89\% $\pm$ 0.73\%} & 93.36\% $\pm$ 1.18\% \\ \cline{3-3}\cline{5-9}
    
    && 0.5 && 84.92\% $\pm$ 3.71\% & 83.10\% $\pm$ 2.79\% & 85.54\% $\pm$ 3.39\% & \textbf{88.21\% $\pm$ 2.17\%} & 84.61\% $\pm$ 3.44\% \\ \cline{3-3}\cline{5-9}
    
    && 0.75 && \textbf{77.55\% $\pm$ 3.63\%} & 76.25\% $\pm$ 4.10\% & 77.55\% $\pm$ 3.64\%
    & 74.68\% $\pm$ 5.31\%  & 57.79\% $\pm$ 12.66\% \\ \cline{3-3}\cline{5-9}
    
    && 0.9 && 59.10\% $\pm$ 8.70\% & 59.02\% $\pm$ 9.19\% & \textbf{59.46\% $\pm$ 8.74\%} & 57.29\% $\pm$ 10.72\%  & 50.51\% $\pm$ 5.57\% \\ \cline{1-9}
    
    \multicolumn{4}{||c|}{Number of times that performed the best} & 4 & 1 & 12 & 2 & 6   \\ \cline{1 -9}

    \end{tabular}}
    \label{tab:FL_all_agg}
\end{table*}

\begin{table*}[htbp]
    \centering
    \caption{\hl{Curvature values derived from the CNN model's accuracy trends at various levels of non-IID data, highlighting points of sharp change in performance under increasing label skew across five datasets.}}
    \resizebox{\textwidth}{!}{
    \begin{tabular}{||c|c|c|c|c|c|c|c|c||}
    \hline
    Dataset & \textsf{HD}=0.25 &\textbf{\textsf{HD}=0.50}
    &\textsf{HD}=0.60
    &\textsf{HD}=0.65
    &\textsf{HD}=0.70
    &\textbf{\textsf{HD}=0.75}
    &\textsf{HD}=0.80
    &\textsf{HD}=0.85\\
    \hline\hline

    CIFAR10 & 0.2 & \textbf{0.3} & 1.1 & 0.5 & 1.6 & \textbf{3.9} & 4.4 & 2.8\\ 
    \hline

    Covtype & 0.3 & \textbf{0.4} & 0.2 & 0.3 & 0.2 & \textbf{3.5} & 6.6 & 2.9\\ 
    \hline

    FMNIST & 0.0 & \textbf{0.0} & 0.5 & 0.6 & 0.9 & \textbf{1.3} & 1 & 3.3 \\ 
    \hline

    Physionet & 0.1 & \textbf{0.2} & 0.5 & 2.2 & 1.3 & \textbf{1} & 3.3 & 3.2 \\ 
    \hline

    CIFAR100 & 0.0 & \textbf{0.2} & 0.1 & 0.1 & 1.2 & \textbf{1.5} & 2.5 & 1.8\\ 
    \hline  
    \hline

    \#Detected as critical point & 0 & \textbf{0} & 1 & 1 & 3 & \textbf{5} & 5 & 5\\ 
    \hline  
     Average curvature & 1.2 & \textbf{0.2} & 0.5 & 0.75 & 1 & \textbf{3.7} & 3.57 & 2.8\\ 
    \hline  
    
    \end{tabular}}
\label{tab:labelskew_allDataset_curvature}
\end{table*}

\begin{table*}[htbp]
    \centering
    \caption{\hl{Curvature values derived from the accuracy trends of CNN, EfficientNetB0, and MobileNetV2 models on the CIFAR-10 dataset at various levels of non-IID data, indicating points of sharp performance change.}}
    \resizebox{\textwidth}{!}{
    \begin{tabular}{||c|c|c|c|c|c|c|c|c||}
    \hline
    Model & \textsf{HD}=0.25 &\textbf{\textsf{HD}=0.50}
    &\textsf{HD}=0.60
    &\textsf{HD}=0.65
    &\textsf{HD}=0.70
    &\textbf{\textsf{HD}=0.75}
    &\textsf{HD}=0.80
    &\textsf{HD}=0.85\\
    \hline\hline

    CNN & 0.2 & \textbf{0.3} & 1.1 & 0.5 & 1.6 & \textbf{3.9} & 4.4 & 2.8 \\ 
    \hline

    EfficientNetB0 & 0.2 & \textbf{0.5} & 2.3 & 3.1 & 4.7 & \textbf{3.7} & 2.1 & 0.7 \\ 
    \hline

    MobileNetV2 & 0.2 & \textbf{0.3} & 2.0 & 2.9 & 5.5 & \textbf{4.8} & 1.1 & 0.3\\ 
    \hline
    \hline

    \#Detected as critical point & 0 & \textbf{0} & 3 & 2 & 3 & \textbf{3} & 3 & 1\\ 
    \hline  
    Average curvature & 0.2 & \textbf{0.4} & 1.8 & 2.2 & 3.9 & \textbf{4.1} & 2.5 & 1.3\\ 
    \hline

    \end{tabular}}
\label{tab:labelskew_cifar10_allmodels_curvature}
\end{table*}

\begin{tcolorbox}[colback=blue!5!white,colframe=blue!75!black, boxsep=0mm, left=2mm, right=2mm, top=1mm, bottom=1mm]
\textbf{Highlight 1}: The drop in the model's performance for label skew appears in a double threshold. A notable performance decline is immediately evident when the \textsf{HD} exceeds 0.5 and 0.75.
\end{tcolorbox}

\begin{figure}[h!]
\includegraphics[width=8.89cm,height=8.89cm,keepaspectratio]{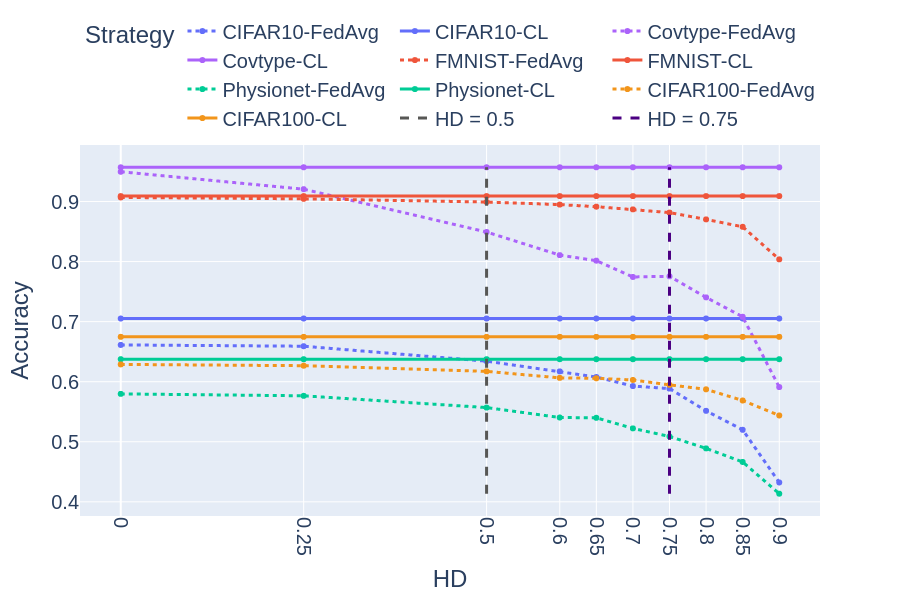}
  \caption{Changes in the models' accuracy considering different levels of \hl{non-IID data} measured by \textsf{HD} for $K = 30$. 
  }
  \label{fig:all_accuracy}
\end{figure}

Previous works claim that \hl{non-IID data} affects the performance of FL models~\cite{lu2024federated, ma2022state, jamali2022federated}. Nevertheless, for the first time, we showcase that the effect of \hl{non-IID data} is not the same under all levels of heterogeneity. Figure~\ref{fig:all_accuracy} depicts the accuracy change by varying the level of \hl{non-IID data} distributions among the clients measured by \textsf{HD} concerning the baseline model created in the centralized setting. As the \hl{non-IID data} partitions increase, the model's accuracy decreases. When the \textsf{HD} between data distributions of the clients exceeds 0.75, the drop becomes more drastic compared to previous levels. 

\hl{To more precisely characterize the inflection points in model performance, we compute the curvature of the accuracy curves across multiple datasets. As shown in Table}~\ref{tab:labelskew_allDataset_curvature}\hl{, curvature values highlight sharper changes at \textsf{HD} = 0.75 and beyond. Notably, models also start to experience a sharper decline in accuracy after \textsf{HD} = 0.5, indicating the beginning of instability. The aggregated metric "\#Detected as critical point" shows that \textsf{HD} = 0.75 is identified as a critical point in all five datasets, and the average curvature at this point (3.7) is the highest across all \textsf{HD} values.}

One possible explanation for this double-threshold effect is that when \textsf{HD} surpasses 0.5, the model starts experiencing a noticeable decline due to increasing divergence in local distributions, leading to a degradation in the global model's generalization. However, beyond \textsf{HD} = 0.75, the level of heterogeneity may reach a critical point where client models become overly specialized to their local data, significantly reducing the effectiveness of global aggregation. This sharp accuracy drop suggests that at extreme levels of \hl{non-IID data}, FedAvg struggles to find a well-generalized solution, potentially due to conflicting optimization directions from highly dissimilar client updates.

\begin{tcolorbox}[colback=blue!5!white,colframe=blue!75!black, boxsep=0mm, left=2mm, right=2mm, top=1mm, bottom=1mm]
\textbf{Highlight 2}: Transfer learning models exhibit greater sensitivity to variations in clients' label distributions, with performance degrading more rapidly and sharply as non-IID data increases.
\end{tcolorbox}

\begin{figure}[h!]
\includegraphics[width=8.89cm,height=8.89cm,keepaspectratio]{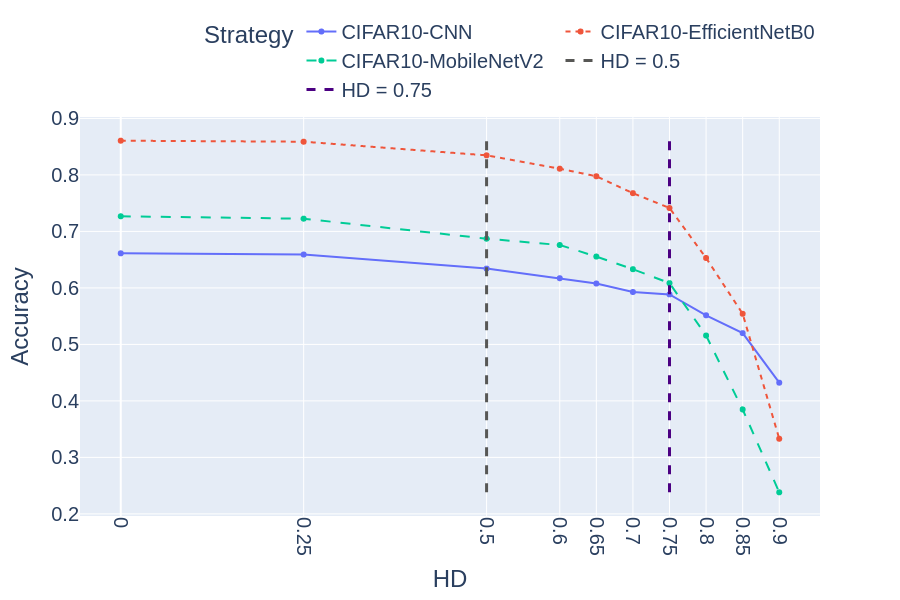}
  \caption{Changes in the models' accuracy considering different levels of \hl{non-IID data} measured by \textsf{HD} for $K = 30$ using CNN, EfficientNetB0, and MobileNetV2 on the CIFAR10 dataset. }
  \label{fig:cifar10_3models_FedAvg}
\end{figure}

\hl{Figure}~\ref{fig:cifar10_3models_FedAvg} \hl{illustrates the accuracy of models created using three different architectures: the CNN discussed earlier, EfficientNetB0, and MobileNetV2, both of which utilize transfer learning. As \textsf{HD} increases, all models experience performance degradation. Transfer learning models exhibit a more pronounced decline under high non-IID data compared to the CNN which stems from their reliance on frozen feature extractors, limiting adaptation to heterogeneous data. As \textsf{HD} increases, local updates to the final layers create misaligned feature representations, reducing the effectiveness of global aggregation. In contrast, the CNN trained from scratch better adapts to decentralized data, making it more robust in extreme non-IID settings.}


\hl{Table}~\ref{tab:labelskew_cifar10_allmodels_curvature} \hl{presents the curvature analysis for the same models. The curvature values for the transfer learning models not only confirm the sharper performance drops but also reveal that these declines begin earlier along the \textsf{HD} spectrum. The peaks in curvature, which quantify the steepness of accuracy degradation, occur sooner and with higher intensity compared to those in Table}~\ref{tab:labelskew_allDataset_curvature}. \hl{This suggests that transfer learning models are more sensitive to label distributional shifts over the clients' data. Still, the presence of distinct inflection points after \textsf{HD} = 0.5 and at more prominent at \textsf{HD} = 0.75 further supports the existence of a double threshold pattern in performance degradation under non-IID conditions.}



\begin{tcolorbox}[colback=blue!5!white,colframe=blue!75!black, boxsep=0mm, left=2mm, right=2mm, top=1mm, bottom=1mm]
\textbf{Highlight 3}: Aggregation algorithms such as Rand, POC, and MOON are particularly vulnerable to performance degradation under conditions of high \hl{non-IID data}.
\end{tcolorbox}

Consider the case when the \textsf{HD} is 0.9 (i.e., high \hl{non-IID data}) for all the datasets presented in Table~\ref{tab:FL_all_agg}. When comparing the accuracy of Rand and POC versus their corresponding CL performance, those algorithms tend to have a higher drop in performance than the other aggregation algorithms. In FL scenarios with high \hl{non-IID data}, the distribution of data labels across clients is highly uneven. It means that specific clients may have more or different data types than others. Under such a scenario, Rand and POC may inadvertently select clients with skewed or unrepresentative data distributions, leading to poor generalization performance when aggregating their updates. 

MOON has the sharpest decrease in performance when the paradigm switches from CL to FL. MOON's contrastive loss, designed to align local and global representations, becomes ineffective when client distributions are too divergent, as past representations no longer serve as meaningful anchors. This misalignment exacerbates performance degradation, making these methods less suited for extreme non-IID scenarios.

\begin{tcolorbox}[colback=blue!5!white,colframe=blue!75!black, boxsep=0mm, left=2mm, right=2mm, top=1mm, bottom=1mm]
\textbf{Highlight 4}: Unbalanced class datasets experience a sharper drop in performance when moving from IID to highly non-IID settings compared to balanced datasets.
\end{tcolorbox}

Consider the IID case (i.e., \textsf{HD} = 0) and the most extreme non-IID case (i.e., \textsf{HD} = 0.9) for each dataset as depicted in Table~\ref{tab:CL_to_FL_datasets}. Notice that the range of decrease in the reported performance (accuracy of \textsf{HD}=0 - accuracy of \textsf{HD}=0.9) is higher for the unbalanced dataset (Covtype) than for balanced datasets (CIFAR10, FMNIST, CIFAR100, and Physionet). 

\begin{table}[htbp]
    \centering
    \caption{The performance's decrease range of the model for the four datasets, moving from the lowest (\textsf{HD}=0) to the highest (\textsf{HD}=0.9) levels of \hl{non-IID data} and considering FedAvg. 
    } 
    \resizebox{0.4\columnwidth}{!}{
    \begin{tabular}{||c|c|c|c|c||}
    \hline
    Dataset & (\textsf{HD}=0) - (\textsf{HD}=0.9) \\
    \hline\hline
    CIFAR10 & 22.9\% \\
    \hline
    FMNIST & 10.31\% \\
    \hline
    CIFAR100 & 8.5\% \\
    \hline
    Physionet & 16.62\% \\
    \hline
    Covtype & 35.85\% \\
    \hline
    \end{tabular}}
    \label{tab:CL_to_FL_datasets}
\end{table}

The sharper performance drop in unbalanced datasets under high \hl{non-IID data} derives from the compounding effects of class imbalance and \hl{non-IID data}. In such cases, certain classes may be overrepresented in specific clients while being nearly absent in others, leading to biased local models. These biased updates fail to capture the overall class distribution when aggregated, resulting in poor generalization. In contrast, balanced datasets distribute class information more evenly across clients, mitigating this effect and leading to a less severe performance decline.

\subsection{Convergence}
In this subsection, we focus on the findings obtained using the CIFAR10 dataset to compare different aggregation algorithms regarding their learning process and the smoothness of training. 

\begin{tcolorbox}[colback=blue!5!white,colframe=blue!75!black, boxsep=0mm, left=2mm, right=2mm, top=1mm, bottom=1mm]
\textbf{Highlight 5}: The higher the \hl{non-IID data} level of labels, the more rounds are required to achieve convergence~\cite{li2022federated}.
\end{tcolorbox}

\begin{table}[htbp]
    \centering
    \caption{RTA for FedAvg, Rand, FedProx, POC, and MOON reached in different levels of non-IID cases as determined by \textsf{HD} over CIFAR10 for K = 30.} 
    \resizebox{\columnwidth}{!}{
    \begin{tabular}{||c|c|c|c|c|c|c|c|c|c||}
    \hline
    Category& Dataset & \makecell{Aggregation \\algorithm} & \textsf{HD} = 0 & \textsf{HD} = 0.25 & \textsf{HD} = 0.5 & \textsf{HD} = 0.75 & \textsf{HD} = 0.9 \\
    \hline\hline

    \multirow[c]{5}{*}{\makecell{Label \\distribution \\skew}} & \multirow[c]{5}{*}{CIFAR10}& FedAvg & 5 & 5 & 6 & 9 & 15 \\ \cline{3-8}

    & & Rand & 5 & 5 & 6 & 10 & 14 \\ \cline{3-8}

    & & FedProx & 5 & 5 & 6 & 9 & 15 \\ \cline{3-8}

    & & POC & 5 & 5 & 6 & 8 & 15 \\ \cline{3-8}

    & & MOON & 8 & 8 & 10 & 12 & 20 \\ \cline{1-8}
    
    \end{tabular}}
    \label{tab:Round_to_max_accuracy}
\end{table}

Table~\ref{tab:Round_to_max_accuracy} examines the algorithms' convergence from a different perspective. We investigate the performance of each aggregation approach across a certain level of non-IID scenario, independent of the others. For each specific non-IID situation described by \textsf{HD}, we determine how many rounds each aggregation algorithm requires to achieve 90\% of its maximum accuracy. Therefore, regardless of the aggregation algorithm, as the \hl{non-IID data} partitioning over the clients increases, more communication rounds are necessary to reach a convergence point (where the accuracy gets stable). Such behavior aligns with the findings of Li et al.~\cite{li2022federated}.

We observe the mentioned behavior because the data distributions among clients are sufficiently dissimilar, so the models built for each client are only optimal for their data, which diverges from the optimal case. As the training progresses, the weights delivered by the server to the clients improve because they get optimized by considering all of the data across all clients.

\begin{table*}[htbp]
    \centering
    \caption{Mean and standard deviation Accuracy for each dataset for CL, FedAvg, Rand, FedProx, Power-Of-Choice, and MOON, considering different levels of feature skewness measured by \textsf{FHD} for K = 30. Each model has undergone ten different trials (random seeds). 
    }
    \resizebox{\textwidth}{!}{
    \begin{tabular}{||c|c|c|c|c|c|c|c|c|c||}
    \hline
    Category& Method &Dataset & \textsf{FHD} & CL & FedAvg & Rand & FedProx & POC & MOON\\
    \hline\hline

    
    \multirow[c]{36}{*}{ \rotatebox[origin=c]{90}{Feature distribution skew}}& \multirow[c]{16}{*}{ \rotatebox[origin=c]{90}{Gaussian Noise}} & \multirow[c]{4}{*}{CIFAR10}& 0 & 70.50\% $\pm$ 0.60\% & 66.12\% $\pm$ 0.70\% & 66.16\% $\pm$ 0.74\%  & \textbf{66.35\% $\pm$ 0.72\%} & 66.26\% $\pm$ 0.70\% & 64.45\% $\pm$ 1.05\%\\ \cline{4-10}

    &&& 0.35 & 70.81\% $\pm$ 0.45\% & 66.18\% $\pm$ 0.57\% & 66.04\% $\pm$ 0.44\% & \textbf{66.29\% $\pm$ 0.63\%} & 66.24\% $\pm$ 0.50\% & 64.23\% $\pm$ 0.53\%  \\ \cline{4-10}
    
    &&& 0.75 & 70.44\% $\pm$ 0.29\% & 66.17\% $\pm$ 0.44\% & 66.31\% $\pm$ 0.61\% & \textbf{66.36\% $\pm$ 0.56\%} & 66.27\% $\pm$ 0.56\%  & 64.58\% $\pm$ 0.55\% \\ \cline{4-10}
    
    &&& 0.9 & 69.71\% $\pm$ 0.03\% & 65.81\% $\pm$ 0.82\% & 65.89\% $\pm$ 0.72\% & 65.85\% $\pm$ 0.68\% & \textbf{65.90\% $\pm$ 0.77\%}  & 63.52\% $\pm$ 0.66\% \\ \cline{3-10}
    
    && \multirow[c]{4}{*}{FMNIST}& 0 & 90.90\% $\pm$ 0.20\% & 90.68\% $\pm$ 0.18\% & 90.57\% $\pm$ 0.14\% & \textbf{90.69\% $\pm$ 0.15\%} & 90.62\% $\pm$ 0.17\% & 88.70\% $\pm$ 0.27\%  \\ \cline{4-10}
    
    &&& 0.35 & 90.91\% $\pm$ 0.26\% & 90.69\% $\pm$ 0.19\% & \textbf{90.97\% $\pm$ 0.15\%} & 90.71\% $\pm$ 0.17\% &  90.97\% $\pm$ 0.17\%  & 88.67\% $\pm$ 0.28\%  \\ \cline{4-10}
    
    &&& 0.75 & 90.96\% $\pm$ 0.11\% & 90.54\% $\pm$ 0.09\% & 90.53\% $\pm$ 0.11\% & \textbf{90.57\% $\pm$ 0.23\%} & 90.51\% $\pm$ 0.19\% & 88.64\% $\pm$ 0.19\%  \\ \cline{4-10}
    
    &&& 0.9 & 90.76\% $\pm$ 0.13\%  & 90.59\% $\pm$ 0.09\%  & \textbf{90.71\% $\pm$ 0.15\%} & 90.70\% $\pm$ 0.29\%  & 90.51\% $\pm$ 0.11\% & 88.55\% $\pm$ 0.21\%   \\ \cline{3-10}

    && \multirow[c]{4}{*}{Physionet}& 0 & 63.74\% $\pm$ 1.24\% & 57.97\% $\pm$ 0.49\% & 57.48\% $\pm$ 0.40\% & 58.16\% $\pm$ 0.62\% & 57.86\% $\pm$ 0.76\%  & \textbf{61.80\% $\pm$ 0.62\%}  \\ \cline{4-10}
    
    &&& 0.35 & 63.30\% $\pm$ 1.31\% & 57.89\% $\pm$ 0.39\%  & 57.92\% $\pm$ 0.64\% & 58.08\% $\pm$ 0.61\% & 57.47\% $\pm$ 1.13\%  & \textbf{61.22\% $\pm$ 0.72\%} \\ \cline{4-10}
    
    &&& 0.75 & 60.13\% $\pm$ 1.11\% & 52.81\% $\pm$ 0.83\%  & 52.84\% $\pm$ 0.74\% & 52.28\% $\pm$ 0.38\% & 51.39\% $\pm$ 0.25\% & \textbf{56.10\% $\pm$ 0.90\%} \\ \cline{4-10}
    
    &&& 0.9 & 28.97\% $\pm$ 3.22\% & 29.43\% $\pm$ 1.49\% & 29.88\% $\pm$ 1.27\% & 29.43\% $\pm$ 1.15\% & 25.25\% $\pm$ 1.97\%  & \textbf{32.06\% $\pm$ 1.30\%}  \\ \cline{3-10}

    && \multirow[c]{4}{*}{Covtype}& 0 & 95.60\% $\pm$ 0.10\% & 94.95\% $\pm$ 0.06\% & 94.89\% $\pm$ 0.08\% & 94.96\% $\pm$ 0.09\% & 94.84\% $\pm$ 0.10\%  & \textbf{95.53\% $\pm$ 0.04\%}  \\ \cline{4-10}
    
    &&& 0.35 & 95.68\% $\pm$ 0.10\%  &  94.94\% $\pm$ 0.06\% & 94.90\% $\pm$ 0.04\%  & 94.90\% $\pm$ 0.08\% & 94.88\% $\pm$ 0.08\%  & \textbf{95.65\% $\pm$ 0.06\%} \\ \cline{4-10}
    
    &&& 0.75 & 95.53\% $\pm$ 0.04\% & 94.79\% $\pm$ 0.08\% & 94.62\% $\pm$ 0.04\%  & 94.74\% $\pm$ 0.07\% & 94.68\% $\pm$ 0.13\% & \textbf{95.11\% $\pm$ 0.07\%} \\ \cline{4-10}
    

    &&& 0.9 & 68.53\% $\pm$ 1.49\%  &  50.01\% $\pm$ 0.35\% & 49.81\% $\pm$ 0.50\%  & 50.03\% $\pm$ 0.42\% & \textbf{50.10\% $\pm$ 1.67\%} & 49.20\% $\pm$ 0.11\% \\ \cline{2-10}
    
    & \multirow[c]{20}{*}{ \rotatebox[origin=c]{90}{Hist-Dirichlet}} & \multirow[c]{5}{*}{CIFAR10}& 0 & \multirow[c]{5}{*}{70.50\% $\pm$ 0.60\%} & \textbf{66.42\% $\pm$ 0.34\%} & 66.35\% $\pm$ 0.35\%  & 66.21\% $\pm$ 0.59\% & 66.40\% $\pm$ 0.70\% & 64.79\% $\pm$ 0.32\% \\ \cline{4-4} \cline{6-10}
    
    &&& 0.25 && 66.09\% $\pm$ 0.49\% & 66.13\% $\pm$ 0.48\% & 65.82\% $\pm$ 0.46\% & \textbf{66.15\% $\pm$ 0.53\%} & 65.12\% $\pm$ 0.85\% \\ \cline{4-4} \cline{6-10}

    &&& 0.5 && 66.30\% $\pm$ 0.67\% & 66.04\% $\pm$ 0.66\% & 66.22\% $\pm$ 0.40\% & \textbf{66.52\% $\pm$ 0.56\%} & 64.52\% $\pm$ 1.08\% \\ \cline{4-4} \cline{6-10}
    
    &&& 0.75 && 66.23\% $\pm$ 0.33\% & 66.04\% $\pm$ 0.67\% & 66.17\% $\pm$ 0.55\% & \textbf{66.34\% $\pm$ 0.51\%}  & 64.99\% $\pm$ 0.31\% \\ \cline{4-4} \cline{6-10}
    
    &&& 0.9 && \textbf{66.25\% $\pm$ 0.47\%} & 65.25\% $\pm$ 0.51\% & 65.25\% $\pm$ 0.90\% & 66.17\% $\pm$ 0.47\%  & 64.12\% $\pm$ 0.56\% \\ \cline{3-10}
    
    && \multirow[c]{5}{*}{FMNIST}& 0 & \multirow[c]{5}{*}{90.90\% $\pm$ 0.20\%} & \textbf{90.72\% $\pm$ 0.20\%}  & 90.68\% $\pm$ 0.12\% & 90.68\% $\pm$ 0.15\% & 90.59\% $\pm$ 0.22\% & 88.75\% $\pm$ 0.13\% \\ \cline{4-4} \cline{6-10}

    &&& 0.25 && 90.62\% $\pm$ 0.10\% & 90.66\% $\pm$ 0.27\% & \textbf{90.70\% $\pm$ 0.15\%} & 90.62\% $\pm$ 0.18\% & 88.72\% $\pm$ 0.32\% \\ \cline{4-4} \cline{6-10}
    
    &&& 0.5 && \textbf{90.78\% $\pm$ 0.12\%} & 90.66\% $\pm$ 0.09\% & 90.63\% $\pm$ 0.17\% & 90.78\% $\pm$ 0.23\% & 88.43\% $\pm$ 0.30\% \\ \cline{4-4} \cline{6-10}
    
    &&& 0.75 && 90.53\% $\pm$ 0.24\% & 90.67\% $\pm$ 0.18\% & \textbf{90.73\% $\pm$ 0.17\%} & 90.56\% $\pm$ 0.12\% & 88.26\% $\pm$ 0.30\% \\ \cline{4-4} \cline{6-10}
    
    &&& 0.9 && 89.77\% $\pm$ 0.32\% & 89.73\% $\pm$ 0.35\% & 89.66\% $\pm$ 0.24\% & \textbf{89.81\% $\pm$ 0.29\%} & 87.38\% $\pm$ 0.16\% \\ \cline{3-10}
    
    && \multirow[c]{5}{*}{Physionet}& 0 & \multirow[c]{5}{*}{63.74\% $\pm$ 1.24\%} & 57.73\% $\pm$ 0.72\% & 57.76\% $\pm$ 0.51\% & 58.02\% $\pm$ 0.76\% & 57.52\% $\pm$ 0.68\% & \textbf{61.13\% $\pm$ 0.42\%}  \\ \cline{4-4} \cline{6-10}

    &&& 0.25 && 57.67\% $\pm$ 0.61\% & 57.62\% $\pm$ 0.69\% & 58.16\% $\pm$ 0.63\% & 57.05\% $\pm$ 0.37\% & \textbf{61.91\% $\pm$ 0.39\%} \\ \cline{4-4} \cline{6-10}
    
    &&& 0.5 && 57.92\% $\pm$ 0.40\% & 57.50\% $\pm$ 0.68\% & 57.86\% $\pm$ 0.33\% & 57.35\% $\pm$ 0.47\% & \textbf{61.20\% $\pm$ 0.83\%} \\ \cline{4-4} \cline{6-10}
    
    &&& 0.75 && 57.27\% $\pm$ 0.64\% & 57.18\% $\pm$ 0.97\% & 57.34\% $\pm$ 0.88\% & 56.99\% $\pm$ 0.86\% & \textbf{62.11\% $\pm$ 0.44\%}\\ \cline{4-4} \cline{6-10}
    
    &&& 0.9 && 56.47\% $\pm$ 0.80\% & 55.49\% $\pm$ 1.34\% & 56.49\% $\pm$ 0.60\% & 55.80\% $\pm$ 1.24\%  & \textbf{59.82\% $\pm$ 1.02\%} \\ \cline{3-10}

    && \multirow[c]{5}{*}{Covtype}& 0 & \multirow[c]{5}{*}{95.60\% $\pm$ 0.10\% } & 94.95\% $\pm$ 0.03\% & 94.81\% $\pm$ 0.02\% & 95.00\% $\pm$ 0.03\% & \textbf{98.84\% $\pm$ 0.09\%}  & 95.62\% $\pm$ 0.11\%  \\ \cline{4-4} \cline{6-10}
    
    &&& 0.25 && 94.95\% $\pm$ 0.05\% & 94.83\% $\pm$ 0.09\% & 94.97\% $\pm$ 0.09\% & 94.77\% $\pm$ 0.03\%  & \textbf{95.63\% $\pm$ 0.05\%} \\ \cline{4-4} \cline{6-10}
    
    &&& 0.5 && 94.90\% $\pm$ 0.02\% & 94.88\% $\pm$ 0.11\% & 94.90\% $\pm$ 0.07\% & 94.88\% $\pm$ 0.02\% & \textbf{95.65\% $\pm$ 0.02\%} \\ \cline{4-4} \cline{6-10}

    &&& 0.75 && 94.80\% $\pm$ 0.05\% & 94.67\% $\pm$ 0.10\% & 94.78\% $\pm$ 0.08\% & 94.74\% $\pm$ 0.10\% & \textbf{95.57\% $\pm$ 0.07\%} \\ \cline{4-4} \cline{6-10}
    
    &&& 0.9 && 93.30\% $\pm$ 0.42\% & 93.11\% $\pm$ 0.52\% & 93.38\% $\pm$ 0.31\% & 93.47\% $\pm$ 0.33\%  & \textbf{94.51\% $\pm$ 0.07\%} \\ \cline{1-10}
    
    \multicolumn{5}{||c|}{Number of times that performed the best} & 4 & 2 & 7 & 7 & 16  \\ \cline{1-10}

    \end{tabular}}
    \label{tab:FL_all_agg_feature_skew}
\end{table*}

Furthermore, FedAvg, Rand, FedProx, and POC exhibit similar behavior in achieving a convergence point, and they do so after roughly the same number of communication rounds. On the other hand, MOON, as expected, requires more rounds to reach the same convergence state.

%% file: 06-section-feature-results.tex
This section examines how feature skew in the client data affects the models' performance. 

\subsection{Synthetic Partitioning Method} 
For simulating feature skew, we employed
two diverse methods from FedArtML~\cite{jimenez2024fedartml} to test their properties:

\emph{Gaussian noise method:} 
This approach introduces diverse Gaussian noise levels to each client's
local dataset to achieve diverse feature distributions. Specifically,
for each client $i$, noise levels $\hat{x}$ are added according to the
user-defined noise level $\sigma$, with $\hat{x} \sim \text{Gau}
\left(\sigma \cdot \frac{i}{K}\right)$, where $\hat{x}$ represents the
resultant features after applying the noise level to the original
features. Here, $\text{Gau}\left(\sigma \cdot \frac{i}{K}\right)$
denotes a Gaussian distribution with a mean of 0 and a variance of
$\sigma \cdot \frac{i}{K}$, and $K$ represents the total number of
clients.

\emph{Hist-Dirichlet-based method:} 
The process starts by characterizing the attributes of each client
using other average values and then subjecting them to a binning
procedure. Subsequently, it establishes the participation of each
feature category within each client using the DD with a specified
$\alpha$. Unlike the Gaussian Noise approach, this method distributes
the data among the clients without modifying the features. We measure the \hl{non-IID data} with the \textsf{HD} among the features across
clients (\textsf{FHD}) within the range $\{0, 0.25, 0.5, 0.75, 0.9\}$.

\subsection{Classification Power} In this subsection, we focus on the findings from the simulations to compare different aggregation techniques and datasets regarding their classification power (a.k.a. accuracy) in the presence of feature skew over the clients' data. Table~\ref{tab:FL_all_agg_feature_skew} summarizes the models' performance derived from different aggregation algorithms under varying degrees of non-IID feature distributions, as indicated by \textsf{FHD}. 

\begin{tcolorbox}[colback=blue!5!white,colframe=blue!75!black, boxsep=0mm, left=2mm, right=2mm, top=1mm, bottom=1mm]
\textbf{Highlight 6}: The performance of FL-generated models is lower than that of CL-generated models.
\end{tcolorbox}

Transitioning the training methodology from CL to FL depicts a decline in the model's performance, notably more pronounced in image datasets (CIFAR10) compared to tabular datasets (Covtype). This outcome is predictable since the model gets trained without access to the complete dataset, and each client optimizes the weights based on its available data. The more significant decrease observed in image datasets stems from the heightened complexity inherent in classification tasks compared to tabular datasets.

\begin{tcolorbox}[colback=blue!5!white,colframe=blue!75!black, boxsep=0mm, left=2mm, right=2mm, top=1mm, bottom=1mm]
\textbf{Highlight 7:} The model's performance in image datasets remains unaffected by increasing the feature \hl{non-IID data}~\cite{li2022federated}.
\end{tcolorbox}
Consider only the image datasets (CIFAR10, FMNIST) and the models generated in FL. Regardless of the aggregation algorithm employed, the performance of the final model remains stable across different levels of feature \hl{non-IID data}, consistently converging to specific values for each aggregation algorithm. This behavior is consistent with the observations reported by Li et al.~\cite{li2022federated}.

Such a pattern arises from the robustness of convolutional layers, which extract spatial features while suppressing minor pixel variations. In the Gaussian-noise method, small perturbations do not significantly alter key patterns, as convolutional filters average out noise. Deeper layers further aggregate features, preserving essential information and minimizing the impact on performance.

\begin{tcolorbox}[colback=blue!5!white,colframe=blue!75!black, boxsep=0mm, left=2mm, right=2mm, top=1mm, bottom=1mm]
\textbf{Highlight 8:} For tabular datasets, using Gaussian noise
levels that exceed \textsf{FHD}=0.9 results in a notable performance decline in
the model, emphasizing the acute dissimilarity among samples.
\end{tcolorbox}

Having tabular datasets (Covtype, Physionet) and using the Hist-Dirichlet approach shows that increasing the degree of feature \hl{non-IID data} does not impact the performance. However, when dealing with Gaussian noise, if we surpass \textsf{FHD}=0.9, there's a noticeable decline in performance. 

This decline occurs because the data becomes highly dissimilar and noisy, making it difficult for the model to extract meaningful patterns. Even in CL, where data is typically more stable, excessive noise disrupts feature learning, reducing the model's generalization ability and leading to performance degradation.

\begin{tcolorbox}[colback=blue!5!white,colframe=blue!75!black, boxsep=0mm, left=2mm, right=2mm, top=1mm, bottom=1mm]
\textbf{Highlight 9:} In scenarios where the features are non-IID partitioned across clients, MOON performs better than all other aggregation algorithms for tabular datasets.
\end{tcolorbox}

Table~\ref{tab:FL_all_agg_feature_skew} validates that no particular algorithm outperforms others in image datasets, as they yield comparable final performance metrics. MOON emerges as the top-performing algorithm in tabular datasets, surpassing all other algorithms. Its performance is nearly equivalent to that of models trained in CL. 

This occurs since MOON can immediately start learning meaningful contrasts between differences in the label using the provided features of the tabular dataset. On the other hand, for images, the model first needs to learn to extract meaningful features from raw pixels before it can start contrasting different object classes effectively. For example, consider the Physionet dataset, which contains features such as age, sex, heart rate, and P-R interval. In that case, each feature has a clear medical interpretation, and the model can directly use the mentioned feature values without learning initial representations. Conversely, for CIFAR10, the model must learn to extract meaningful features from raw pixels before contrastive learning can be effective.

\subsection{Convergence} In this subsection, we concentrate on the results of the simulations, aiming to contrast various aggregation algorithms and datasets in terms of their convergence. 

\begin{tcolorbox}[colback=blue!5!white,colframe=blue!75!black, boxsep=0mm, left=2mm, right=2mm, top=1mm, bottom=1mm]
\textbf{Highlight 10:} Feature skew doesn't alter the model convergence point.
\end{tcolorbox}

\begin{table}[htbp]
    \centering
    \caption{RTA for FedAvg, Rand, FedProx, POC, and MOON reached in different levels of feature non-IID cases as determined by \textsf{FHD} over CIFAR10  and Covtype using Hist Dirichlet for K = 30} 
    \resizebox{\columnwidth}{!}{
    \begin{tabular}{||c|c|c|c|c|c|c|c|c|c|c||}
    \hline
    Category& Method & Dataset & \makecell{Aggregation \\algorithm} & \textsf{FHD} = 0 & \textsf{FHD} = 0.25 & \textsf{FHD} = 0.50 & \textsf{FHD} = 0.75 & \textsf{FHD} = 0.9\\
    \hline\hline

    \multirow[c]{10}{*}{\makecell{Feature \\distribution \\skew}} & \multirow[c]{10}{*}{\makecell{Hist \\Dirichlet}} & \multirow[c]{5}{*}{CIFAR10} & FedAvg & 5 & 5 & 5 & 5 & 4  \\ \cline{4-9}
    &&& Rand & 5 & 5 & 5 & 5 & 4 \\ \cline{4-9}
    &&& FedProx & 5 & 5 & 5 & 5 & 4  \\ \cline{4-9}
    &&& POC & 5 & 5 & 5 & 5 & 4  \\ \cline{4-9}
    &&& MOON & 8 & 7 & 7 & 8 & 7 \\ \cline{3-9}   

    && \multirow[c]{5}{*}{Covtype} & FedAvg & 3 & 3 & 3 & 3 & 4  \\ \cline{4-9}
    &&& Rand & 3 & 3 & 3 & 3 & 4 \\ \cline{4-9}
    &&& FedProx & 3 & 3 & 3 & 3 & 4  \\ \cline{4-9}
    &&& POC & 3 & 3 & 3 & 3 & 4 \\ \cline{4-9}
    &&& MOON & 4 & 4 & 4 & 4 & 5  \\ \cline{1-9}   
    \end{tabular}}
\label{tab:Round_to_max_accuracy_FeatureSkew_HD}
\end{table}

Table~\ref{tab:Round_to_max_accuracy_FeatureSkew_HD} presents an alternative viewpoint on the previous highlight. It outlines the iterations needed for FedAvg, Rand, FedProx, POC, and MOON to achieve 90\% of the maximum accuracy across various degrees of feature non-IID conditions, as characterized by \textsf{FHD}, using the CIFAR10 dataset. Increasing the \hl{non-IID data} of features within the data has minimal impact on the model's ability to converge to its optimal performance.

The minimal impact of feature skew on convergence suggests that while feature distributions differ across clients, the underlying task remains learnable. Unlike label skew, which directly affects class representation in local updates, feature skew primarily alters input variations without disrupting the overall decision boundary. As a result, the global model can still generalize effectively across clients, leading to similar convergence behavior regardless of the degree of feature \hl{non-IID data}.

%% file: 07-section-quantity-results.tex
This section examines how the quantity skew in the client data affects the models' performance. 

\subsection{Synthetic Partitioning Method} 
We use the MinSize-Dirichlet method included in the FedArtML~\cite{jimenez2024fedartml} tool, which specifies the DD's $\alpha$ value and generates the desired participation proportions for each client. Subsequently, a minimum
required size, referred to as ''the minimum number of examples,'' is established for
each client. Thus, the minimum proportion size, denoted as $MinSize$, is
calculated as $\textit{MinSize} = \frac{\textit{MinRequiredSize}}{n}$, where $n$
represents the total number of examples in the centralized dataset. If
the designated proportions fall below $\textit{MinSize}$, it substitutes them
with $\textit{MinSize}$. Finally, the proportions are normalized to fall from 0 to 1. 

We assess the level of \hl{non-IID data} using the \textsf{HD} for quantity skew (\textsf{QHD}) within the range $\{0, 0.10,0.17\}$. This small range arises because the finite size of the dataset constrains quantity skew. Unlike other skews, the proportions derived from the quantity distribution cannot exhibit extreme divergence, as the total number of samples restricts how unevenly clients can receive data.

\begin{table*}[htbp]
    \centering
    \caption{Mean and standard deviation Accuracy for each dataset for CL, FedAvg, Rand, FedProx, Power-Of-Choice, and MOON, considering different levels of non-IID partitioning of the record quantities measured by \textsf{QHD} for $K = 30$. Each model has undergone five different trials.} 
    
    \resizebox{\textwidth}{!}{
    \begin{tabular}{||c|c|c|c|c|c|c|c|c|c||}
    \hline
    Category& Method & Dataset & CL & \textsf{QHD} & FedAvg & Rand & FedProx & POC & MOON\\
    \hline\hline

    \multirow[c]{12}{*}{ \rotatebox[origin=c]{90}{Quantity distribution skew}}& \multirow[c]{12}{*}{ \rotatebox[origin=c]{90}{Min-size Dirichlet}} & \multirow[c]{3}{*}{CIFAR10}& \multirow[c]{3}{*}{70.50\% $\pm$ 0.6\%} & 0 & 65.77\% $\pm$ 0.57\% & 65.92\% $\pm$ 0.72\%  & \textbf{66.45\% $\pm$ 0.61\%} & 66.04\% $\pm$ 0.35\% & 64.01\% $\pm$ 0.46\% \\ \cline{5-10}

    &&&& 0.10 & 66.69\% $\pm$ 0.72\% & 66.29\% $\pm$ 0.67\% & 66.71\% $\pm$ 0.76\% & \textbf{68.82\% $\pm$ 0.89\%} & 63.24\% $\pm$ 0.37\%  \\ \cline{5-10}
    
    &&&& 0.17 & 66.04\% $\pm$ 0.65\% & 67.91\% $\pm$ 0.46\% & 68.07\% $\pm$ 0.42\% & \textbf{68.44\% $\pm$ 0.51\%} & 63.49\% $\pm$ 1.25\% \\ \cline{3-10}
    
    && \multirow[c]{3}{*}{FMNIST}& \multirow[c]{3}{*}{90.90\% $\pm$ 0.02\%} & 0 & \textbf{90.72\% $\pm$ 0.23\%} & 90.69\% $\pm$ 0.25\% & 90.64\% $\pm$ 0.15\% & 90.67\% $\pm$ 0.11\%  & 88.49\% $\pm$ 0.24\%  \\ \cline{5-10}
    
    &&&& 0.10 & 90.37\% $\pm$ 0.16\% & 90.39\% $\pm$ 0.22\% & 90.30\% $\pm$ 0.45\% & \textbf{90.78\% $\pm$ 0.13\%}  & 88.04\% $\pm$ 0.27\% \\ \cline{5-10}
    
    &&&& 0.17 & 90.39\% $\pm$ 0.32\% & 90.43\% $\pm$ 0.19\% & 90.44\% $\pm$ 0.27\% & \textbf{90.65\% $\pm$ 0.31\%} & 88.10\% $\pm$ 0.26\%\\ \cline{3-10}
    
    && \multirow[c]{3}{*}{Physionet}& \multirow[c]{3}{*}{63.74\% $\pm$ 1.24\%} & 0 & 58.23\% $\pm$ 0.68\% & 57.13\% $\pm$ 0.56\% & 58.04\% $\pm$ 0.29\% & 57.08\% $\pm$ 0.53\%  & \textbf{61.29\% $\pm$ 0.81\%} \\ \cline{5-10}
    
    &&&& 0.10 & 59.48\% $\pm$ 2.09\% & 59.06\% $\pm$ 1.71\% & 59.42\% $\pm$ 1.23\% & \textbf{61.29\% $\pm$ 0.50\%} & 58.67\% $\pm$ 1.97\%  \\ \cline{5-10}
    
    &&&& 0.17 & 64.22\% $\pm$ 1.27\% & 64.40\% $\pm$ 0.55\% & \textbf{64.92\% $\pm$ 0.71\%} & 64.82\% $\pm$ 0.85\% & 63.86\% $\pm$ 0.40\%  \\ \cline{3-10}
    
    && \multirow[c]{3}{*}{Covtype}& \multirow[c]{3}{*}{95.60\% $\pm$ 0.1\%} & 0 &  94.97\% $\pm$ 0.04\% & 94.87\% $\pm$ 0.06\% & 94.96\% $\pm$ 0.07\% & 94.83\% $\pm$ 0.05\% & \textbf{95.15\% $\pm$ 0.09\%} \\ \cline{5-10}
    
    &&&& 0.10 & 95.67\% $\pm$ 0.10\%  & 95.65\% $\pm$ 0.07\% & 95.70\% $\pm$ 0.10\% & \textbf{95.79\% $\pm$ 0.10\%} & 95.13\% $\pm$ 0.12\%\\ \cline{5-10}
    
    &&&& 0.17 & 90.65\% $\pm$ 1.57\%  & 94.77\% $\pm$ 0.43\% & \textbf{95.27\% $\pm$ 0.20\%} & 94.94\% $\pm$ 0.44\% & 94.23\% $\pm$ 0.47\% \\ \cline{1-10}
    
    \multicolumn{5}{||c|}{Number of times that performed the best} & 1 & 0 & 3 & 6 & 2  \\ \cline{1-10}

    \end{tabular}}
    \label{tab:FL_all_agg_quantity_skew}
\end{table*}

\subsection{Classification Power: } 

In the following paragraphs, we concentrate on the simulation results to evaluate various aggregation algorithms and datasets regarding their classification accuracy, particularly considering the impact of quantity skew on the clients' data. 

\begin{tcolorbox}[colback=blue!5!white,colframe=blue!75!black, boxsep=0mm, left=2mm, right=2mm, top=1mm, bottom=1mm]
\textbf{Highlight 11:} The quantity skew in the client's data does not affect the performance of the final model~\cite{li2022federated}.
\end{tcolorbox}
\hl{Table}~\ref{tab:FL_all_agg_quantity_skew} \hl{depicts the performance of each aggregation algorithm and dataset, using various levels of non-IID for quantity skew. Examining this table} and considering each aggregation algorithm separately, it is evident that the performance of the final models remains consistent across various levels of quantity skewness in the clients' records. This phenomenon occurs regardless of the chosen aggregation algorithm, confirming that quantity skewness does not affect model performance. This behavior pattern agrees with the results documented by Li et al.~\cite{li2022federated}.

The invariance of model performance to quantity skew suggests that FL aggregation algorithms effectively balance updates regardless of varying client sample sizes. Since clients contribute proportionally to the global model, those with fewer samples still provide functional gradients without disproportionately influencing training. Additionally, standard optimization techniques, such as weighted averaging, mitigate potential biases from data imbalance, ensuring stable performance across different levels of quantity skewness.

\subsection{Convergence}
In this subsection, we focus on the findings obtained when different aggregation algorithms are considered regarding their learning process and the smoothness of training. 

\begin{tcolorbox}[colback=blue!5!white,colframe=blue!75!black, boxsep=0mm, left=2mm, right=2mm, top=1mm, bottom=1mm]
\textbf{Highlight 12:} All aggregation algorithms converge after the same number of communication rounds in the presence of quantity skew.
\end{tcolorbox}

\begin{table}[htbp]
    \centering
    \caption{RTA for FedAvg, Rand, FedProx, POC, and MOON reached in different levels of quantity non-IID cases as determined by \textsf{QHD} over CIFAR10 and Covtype using Min-size Dirichlet method for $K = 30$
    }
    \resizebox{\columnwidth}{!}{
    \begin{tabular}{||c|c|c|c|c|c|c||}
    \hline
    Category& Method & Dataset & \makecell{Aggregation \\algorithm } & \textsf{QHD} = 0 & \textsf{QHD} = 0.10 & \textsf{QHD} = 0.17 \\
    \hline\hline

    \multirow[c]{10}{*}{\makecell{Quantity \\distribution \\skew}} & \multirow[c]{10}{*}{\makecell{Min-size\\ Dirichlet}} & \multirow[c]{5}{*}{CIFAR10} & FedAvg & 5 & 3 & 2 \\ \cline{4-7}
    &&& Rand & 5 & 3 & 1 \\ \cline{4-7}
    &&& FedProx & 5 & 3 & 2 \\ \cline{4-7}
    &&& POC & 5 & 3 & 2 \\ \cline{4-7}
    &&& MOON & 6 & 3 & 2 \\ \cline{3-7}   
    && \multirow[c]{5}{*}{Covtype} & FedAvg & 3 & 2 & 1  \\ \cline{4-7}
    &&& Rand & 3 & 2 & 1 \\ \cline{4-7}
    &&& FedProx & 3 & 2 & 1 \\ \cline{4-7}
    &&& POC & 3 & 2 & 1 \\ \cline{4-7}
    &&& MOON & 4 & 2 & 1 \\ \cline{1-7}   
    \end{tabular}}
\label{tab:Round_to_max_accuracy_QuantitySkew}
\end{table}

\hl{Table}~\ref{tab:Round_to_max_accuracy_QuantitySkew} \hl{shows the RTA for the aggregation algorithms on various levels of quantity non-IID cases for the analyzed datasets. Looking at this table}, it is clear that regardless of the degree of \hl{non-IID data} in the number of records from each label across clients, all aggregation algorithms converge after a consistent number of communication rounds. 

The consistent convergence across aggregation algorithms stems from the redundancy in client datasets, where each client's data mirrors the overall distribution. This redundancy allows the global model to learn similar patterns from any subset of clients, ensuring that convergence remains stable regardless of quantity skew.

%% file: 08-section-spatemp-results.tex

This section examines how varying levels of data disparity among clients, based on time and location, impact model performance.

\subsection{Synthetic Partitioning Method} 
\begin{figure}[ht]
\centering
\includegraphics[width=0.99\columnwidth]{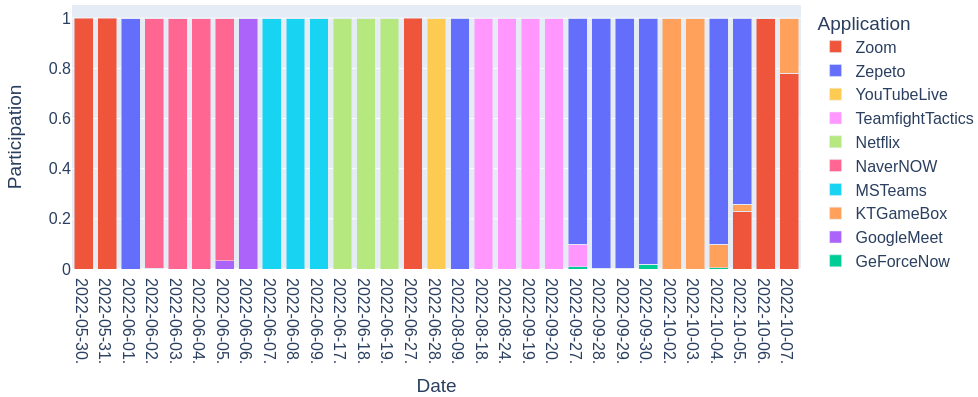}
\caption{Distribution of 5GNTF applications (label) along date (spatiotemporal variable expressed in YYYY-MM-DD).}
\label{fig:example_5g_data_time_label_distro}
\end{figure}

The primary constraint in this partition process is that the dataset must contain a categorical variable of space (i.e., places, cities, latitude, longitude, etc.) or time (i.e., hours, months, years, etc.) to use as the partitioning variable. For instance, Figure~\ref{fig:example_5g_data_time_label_distro} depicts the distribution of labels along the date of the 5GNTF dataset. In this case, the space variable employed to create the federated data is the flow's \emph{date} expressed in year-month-day format (categorical).

We use the St-Dirichlet method from FedArtML~\cite{jimenez2024fedartml}, which employs the DD to segment the data based on spatial (SP skew) or temporal (TMP skew) categories to distribute the data among federated clients. We assess the level of \hl{non-IID data} using the \textsf{HD} for spatiotemporal skew (\textsf{STHD}) within the range \{0, 0.25, 0.5, 0.75, 0.9\}.

\subsection{Classification Power} 
In this subsection, we focus on the simulation results to evaluate various aggregation algorithms and datasets regarding classification accuracy. We pay particular attention to the impact of different levels of spatiotemporal skewness among the clients' data. 

\begin{figure}[h!]
\includegraphics[width=8.89cm,height=8.89cm,keepaspectratio]{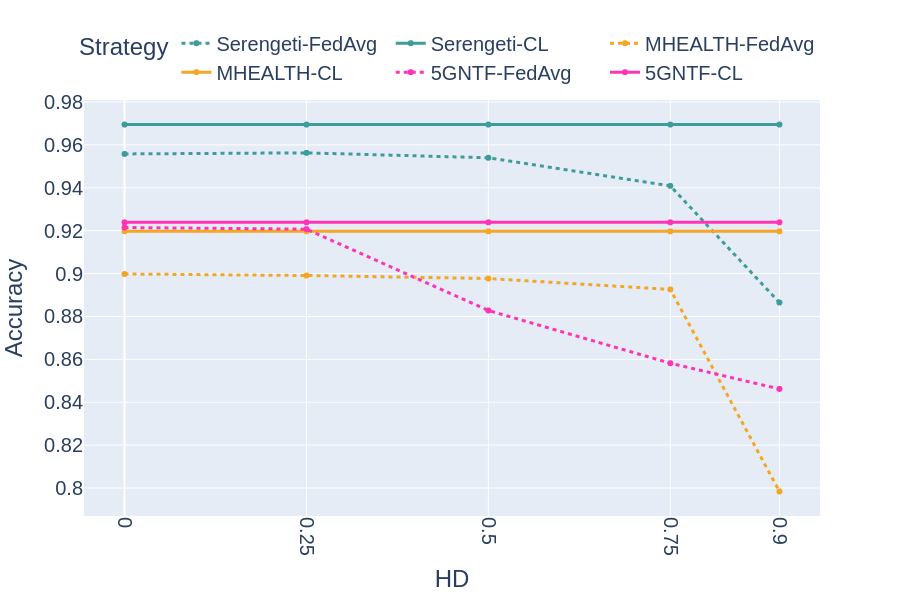}
  \caption{Changes in the models' accuracy considering different levels of \hl{non-IID data} measured by \textsf{STHD} for $K = 30$.}
  \label{fig:all_accuracy_spatiotemp}
\end{figure}

\begin{tcolorbox}[colback=blue!5!white,colframe=blue!75!black, boxsep=0mm, left=2mm, right=2mm, top=1mm, bottom=1mm]
\textbf{Highlight 13:} The higher the \hl{non-IID data} level in time and space, the worse the model’s performance.
\end{tcolorbox}

\begin{table*}[htbp]
    \centering
    \caption{Mean and standard deviation Accuracy for each dataset for CL, FedAvg, Rand, FedProx, Power-Of-Choice, and MOON, considering different levels of non-IID partitioning of the records based on their space (SP) and time (TMP) measured by \textsf{STHD} for $K = 30$. Each model has undergone five trials.} 
    
    \resizebox{\textwidth}{!}{
    \begin{tabular}{||c|c|c|c|c|c|c|c|c|c|c||}
    \hline
    Category& Type & Method & Dataset & CL & \textsf{STHD} & FedAvg & Rand & FedProx & POC & MOON\\
    \hline\hline

    \multirow[c]{15}{*}{ \rotatebox[origin=c]{90}{SPT distribution skew}}& 
    \multirow[c]{10}{*}{ \rotatebox[origin=c]{90}{SP skew}}&\multirow[c]{15}{*}{ \rotatebox[origin=c]{90}{St-Dirichlet}} & \multirow[c]{5}{*}{Serengeti}& \multirow[c]{5}{*}{96.95\% $\pm$ 0.11\%} & 0 & 95.58\% $\pm$ 0.08\% & 95.56\% $\pm$ 0.09\% & 95.62\% $\pm$ 0.14 & 95.48\% $\pm$ 0.09 & \textbf{96.69\% $\pm$ 0.06} \\ \cline{6-11}

    &&&&& 0.25 & 95.63\% $\pm$ 0.05\% & 95.50\% $\pm$ 0.10\% & 95.64\% $\pm$ 0.03 & 95.51\% $\pm$ 0.11 & \textbf{96.76\% $\pm$ 0.08} \\ \cline{6-11}
    
    &&&&& 0.5 & 95.40\% $\pm$ 0.08\% & 95.32\% $\pm$ 0.07\% & 95.36\% $\pm$ 0.06 & 95.32\% $\pm$ 0.06 & \textbf{96.51\% $\pm$ 0.04} \\ \cline{6-11}
    
    &&&&& 0.75 & 94.09\% $\pm$ 0.25\% & 93.91\% $\pm$ 0.36\% & 94.15\% $\pm$ 0.21 & 94.06\% $\pm$ 0.09 & \textbf{95.80\% $\pm$ 0.16} \\ \cline{6-11}
    
    &&&&& 0.9 & 88.65\% $\pm$ 0.18\% & 87.68\% $\pm$ 0.58\% & 88.49\% $\pm$ 0.13 & 88.20\% $\pm$ 0.39 & \textbf{91.74\% $\pm$ 0.20}\\\cline{4-11} 

    &&&\multirow[c]{5}{*}{MHEALTH}& \multirow[c]{5}{*}{91.97\% $\pm$ 0.09} & 0 & 89.98\% $\pm$ 0.04\% & 90.33\% $\pm$ 0.03 & \textbf{90.85\% $\pm$ 0.04} & 90.33\% $\pm$ 0.05 & 90.56\% $\pm$ 0.02 \\ \cline{6-11}
    
    &&&&& 0.25  & 89.91\% $\pm$ 0.05 & 90.34\% $\pm$ 0.04 & \textbf{90.85\% $\pm$ 0.01} & 90.33\% $\pm$ 0.05 & 90.57\% $\pm$ 0.02 \\ \cline{6-11}
    
    &&&&& 0.5  & 89.77\% $\pm$ 0.10 & 90.28\% $\pm$ 0.04 & \textbf{90.85\% $\pm$ 0.02} & 90.32\% $\pm$ 0.04 &  90.48\% $\pm$ 0.04\\ \cline{6-11}
    
    &&&&& 0.75  & 89.26\% $\pm$ 0.31 & 89.15\% $\pm$ 0.24 & \textbf{89.87\% $\pm$ 0.31} & 88.79\% $\pm$ 0.48 & 89.69\% $\pm$ 0.18 \\ \cline{6-11}
    
    &&&&& 0.90  & 79.84\% $\pm$ 0.57 & 82.34\% $\pm$ 1.36 & 82.48\% $\pm$ 0.78 & 81.33\% $\pm$ 0.74 & \textbf{82.95\% $\pm$ 1.04} \\ \cline{2-2}\cline{4-11}


    
    
    

    &\multirow[c]{5}{*}{ \rotatebox[origin=c]{90}{TMP skew}}&& \multirow[c]{5}{*}{5GNTF} & \multirow[c]{5}{*}{92.39\% $\pm$ 0.10\%} & 0 & 92.15\% $\pm$ 0.04\% & 92.15\% $\pm$ 0.02\% & 92.14\% $\pm$ 0.02\% & 92.15\% $\pm$ 0.04\% & \textbf{92.23\% $\pm$ 0.02\%}  \\ \cline{6-11}

    &&&&& 0.25 & 92.07\% $\pm$ 0.12\% & 92.05\% $\pm$ 0.15\% & 92.07\% $\pm$ 0.18\% & 92.15\% $\pm$ 0.05\% & \textbf{92.28\% $\pm$ 0.04\%}  \\ \cline{6-11}
    
    &&&&& 0.5 & 88.28\% $\pm$ 1.87\% & 87.92\% $\pm$ 2.00\% & 89.19\% $\pm$ 2.09\% & \textbf{92.08\% $\pm$ 0.08\%} & 89.24\% $\pm$ 1.80\%  \\ \cline{6-11}
    
    &&&&& 0.75 & 85.82\% $\pm$ 0.06\% & 85.64\% $\pm$ 0.08\% & 85.83\% $\pm$ 0.08\% & \textbf{91.77\% $\pm$ 0.32\%} & 86.51\% $\pm$ 1.79\%  \\ \cline{6-11}
    
    &&&&& 0.9 & 84.62\% $\pm$ 0.36\% & 84.50\% $\pm$ 0.20\% & 84.55\% $\pm$ 0.37\% & \textbf{89.23\% $\pm$ 2.13\%} & 83.94\% $\pm$ 0.02\%  \\ \cline{1-11}

    \multicolumn{6}{||c|}{Number of times that performed the best} & 0 & 0 & 4 & 3 & 8   \\ \cline{1-11}

    \end{tabular}}
    \label{tab:FL_all_agg_spatiotemporal_skew}
\end{table*}

\hl{Table}~\ref{tab:FL_all_agg_spatiotemporal_skew}\hl{ depicts the performance for each dataset and aggregation algorithms, using multiple levels of non-IID spatiotemporal skew. It demonstrates that}, irrespective of the aggregation algorithm employed, model performance deteriorates when data distribution among clients varies concerning time or space. Figure \ref{fig:all_accuracy_spatiotemp} illustrates the comparison between FedAvg and the centralized model across varying STHD levels, showing the same pattern: model accuracy deteriorates as spatiotemporal \hl{non-IID data} among clients grows. However, the magnitude of this deterioration differs across datasets. The performance in all datasets drops noticeably once the STHD surpasses 0.75. This phenomenon occurs because increasing the differences among clients' data based on time and location also raises \hl{non-IID data} in the clients' label distributions. This behavior is evident in Table~\ref{tab:SpatioTemporalSkew_HD}, which displays the \textsf{HD} in label distributions at varying levels of \textsf{STHD} among clients. We also concluded in the label skew study section that higher levels of \hl{non-IID data} among clients' data distributions negatively impact the performance of the final model.

\begin{table}[htbp]
    \centering
    \caption{\textsf{HD} among clients' label distributions at varying levels of non-IID partitioning by time and space} 
    \resizebox{\columnwidth}{!}{
    \begin{tabular}{||c|c|c|c|c|c||}
    \hline
    Dataset &  \textsf{STHD} = 0 & \textsf{STHD} = 0.25 & \textsf{STHD} = 0.50 & \textsf{STHD} = 0.75 & \textsf{STHD} = 0.9\\
    \hline\hline
    Serengeti & 0.01 & 0.09 & 0.22 & 0.36 & 0.53 \\ \cline{1  - 6}
    MHEALTH & 0.01 & 0.01 & 0.01 & 0.03 & 0.07 \\ \cline{1  - 6}
    5GNTF & 0.03 & 0.20 & 0.29 & 0.30  & 0.49 \\ \cline{1 - 6} 
    \end{tabular}
    }
\label{tab:SpatioTemporalSkew_HD}
\end{table}

\subsection{Convergence}
In this subsection, we showcase the results related to the models' convergence when there are variations in the data concerning time and location, considering the results obtained on the Serengeti, MHEALTH, and 5GNTF datasets.

\begin{table}[htbp]
    \centering
    \caption{RTA reached for different levels of spatiotemporal non-IID cases as determined by \textsf{STHD} for K = 30. }
    \resizebox{\columnwidth}{!}{
    \begin{tabular}{||c|c|c|c|c|c|c|c|c||}
    \hline
    Dataset & \makecell{Aggregation \\algorithm} & \textsf{STHD} = 0 & \textsf{STHD} = 0.25 & \textsf{STHD} = 0.50 & \textsf{STHD} = 0.75 & \textsf{STHD} = 0.9\\
    \hline\hline

    \multirow[c]{5}{*}{Serengeti} & FedAvg & 6 & 7 & 7 & 10 & 14  \\ \cline{2-7}
    & Rand & 7 & 7 & 8 & 10 & 14 \\ \cline{2-7}
    & FedProx & 7 & 7 & 7 & 10 & 14  \\ \cline{2-7}
    & POC & 7 & 7 & 7 & 10 & 15  \\ \cline{2-7}
    & MOON & 8 & 8 & 9 & 12 & 19 \\ \cline{1-7}   

    \multirow[c]{5}{*}{5GNTF} & FedAvg & 1 & 1 & 1 & 1 & 1  \\ \cline{2-7}
     & Rand & 1 & 1 & 1 & 1 & 1 \\ \cline{2-7}
     & FedProx & 1 & 1 & 1 & 1 & 1 \\ \cline{2-7}
     & POC & 1 & 1 & 1 & 1 & 1 \\ \cline{2-7}
     & MOON & 1 & 1 & 1 & 1 & 1  \\ \cline{1-7}   

    \multirow[c]{5}{*}{MHEALTH} & FedAvg & 2 & 2 & 2 & 3 & 2  \\ \cline{2-7}
    & Rand & 2 & 2 & 2 & 3 & 2 \\ \cline{2-7}
    & FedProx & 2 & 2 & 2 & 3 & 2 \\ \cline{2-7}
    & POC & 2 & 2 & 2 & 4 & 2 \\ \cline{2-7}
     & MOON & 6 & 7 & 7 & 13 & 14  \\ \cline{1-7}   
    \end{tabular}}
\label{tab:Round_to_max_accuracy_SpatioTemporalSkew}
\end{table}

\begin{tcolorbox}[colback=blue!5!white,colframe=blue!75!black, boxsep=0mm, left=2mm, right=2mm, top=1mm, bottom=1mm] 
\textbf{Highlight 14:} Spatial \hl{non-IID data} shows no consistent impact on convergence rounds; effects vary by dataset dynamics and task difficulty.
\end{tcolorbox}
 
According to Table~\ref{tab:Round_to_max_accuracy_SpatioTemporalSkew}, the same level of \hl{non-IID data} can have a radically different effect depending on the underlying data:

\begin{itemize}
\item \textbf{Serengeti}: As STHD increases from 0 to 0.90, RTA nearly doubles across all aggregation algorithms (e.g., FedAvg: 6 $\rightarrow$ 14; POC: 7 $\rightarrow$ 15). This suggests that data from different sites becomes more heterogeneous, requiring more training rounds for the models to converge.

\item \textbf{5GNTF}: All aggregation algorithms converge in just one round, despite the STHD. This occurs when the classes are very easy to distinguish and there’s a clear gap between the records of one class and those of the others. In such cases, the task becomes trivial, and the time factor has minimal impact on when convergence is reached.

\item \textbf{MHEALTH}: Across all aggregation algorithms, RTA stays constant—except for MOON, which jumps sharply from 6 to 14 rounds as spatiotemporal \hl{non-IID data} goes from 0 to 0.90. Since the data come from body-worn sensors and are split by individual subjects, client-specific covariate shifts emerge that particularly undermine representation-based approaches like MOON.
\end{itemize}

\noindent For the MOON algorithm, the gap in RTA between STHD = 0 and STHD = 0.9 varies by dataset—0 rounds for 5GNTF, 8 for MHEALTH, and 11 for Serengeti—while other aggregation algorithms show no such change. This indicates there is not a one‐to‐one relationship between STHD and convergence speed; instead, factors like task complexity, temporal patterns, and feature diversity shape the outcome, supporting our earlier point that spatial \hl{non-IID data} impacts convergence inconsistently, depending on dataset dynamics and problem difficulty.

%% file: 09-section-general-results.tex
In this section, we provide highlights summarizing the overall results obtained from our experiments, combining the behavior shown before for label, feature, quantity, and spatiotemporal skews.

\begin{tcolorbox}[colback=blue!5!white,colframe=blue!75!black, boxsep=0mm, left=2mm, right=2mm, top=1mm, bottom=1mm]
\textbf{Highlight 15:} Label skew~\cite{vahidian2023rethinking, li2022federated} and spatiotemporal skew significantly impact the model's performance.
\end{tcolorbox}

Our experimental analysis reveals that not all forms of non-IID data equally degrade FL performance. Label skew and spatiotemporal skew exhibit the most severe impact. Label skew reduces model accuracy by 10–40\% compared to the CL baseline. Feature and quantity skews show less significant effects (1-5\% accuracy drops). The previous aligns with prior findings \cite{vahidian2023rethinking, li2022federated} that label skew disproportionately harms aggregation, as local models overfit to dominant classes. 

Spatiotemporal skew introduces contextual drift (e.g., sensor data varying across locations/times), corrupting the feature space. Similarly to label skew, this cannot be fixed through simple aggregation - our tests show FedAvg suffers 10-12\% higher accuracy loss. The global model fails to perform effectively across all contexts because it averages away crucial environmental patterns unique to specific locations or times.

\begin{tcolorbox}[colback=blue!5!white,colframe=blue!75!black, boxsep=0mm, left=2mm, right=2mm, top=1mm, bottom=1mm]
\textbf{Highlight 16:} \hl{FedProx performs better under label skew, POC excels in handling quantity skew, and MOON exhibits greater robustness to feature skew, whereas FedAvg and Rand tend to struggle under high non-IID data levels.}
\end{tcolorbox}

\begin{table}[htbp]
    \centering
    \caption{The number of cases in which each specific algorithm achieved the best performance for each study.} 
    \resizebox{\columnwidth}{!}{
    \begin{tabular}{||c|c|c|c|c|c|c||}
    \hline
    Study &  \#Cases & FedAvg & Rand & FedProx & POC & MOON\\
    \hline\hline
    Label Skew & 25 & 4 & 1 & 12 & 2 & 6 \\ \cline{1  - 7}
    Feature Skew & 36 & 4 & 2 & 7 & 7 & 16 \\ \cline{1  - 7}
    Quantity Skew & 12 & 1 & 0 & 3 & 6 & 2 \\ \cline{1  - 7}
    Spatio Temporal Skew & 15 & 0 & 0 & 4 & 3 & 8 \\ \cline{1  - 7}
    \cline{1  - 7}
    \multicolumn{2}{||c|}{Total best performance} & 9 & 3 & 26 & 18 & 32 \\ 
\cline{ 1 - 7}

    \end{tabular}
    }
\label{tab:Bestperformance_dist_case}
\end{table}

Table~\ref{tab:Bestperformance_dist_case} summarizes the cases in which each specific algorithm exhibited the best performance compared to other aggregation algorithms across four skewness types considered in our study. In most cases, the FedProx, POC, and MOON aggregation algorithms achieved the best performance, outperforming the simpler FedAvg and Rand algorithms.

Such a superior performance can be attributed to the specific mechanisms to tackle the \hl{non-IID data} of each aggregation algorithm. FedProx stabilizes training by regulating the influence of the global model on local clients, POC enhances personalization through loss-based selection, and MOON leverages contrastive learning to improve feature representation. These mechanisms enable better adaptation to diverse client distributions, leading to consistently stronger performance across different skewness types.

Although FedProx, POC, and MOON generally outperform FedAvg and Rand, the performance gains are often marginal. This indicates that while these methods offer improvements in handling non-IID data, they do not fully resolve the challenges posed by \hl{non-IID data}. The relatively small advantage suggests the need for more effective aggregation algorithms to better adapt to diverse client distributions and enhance model performance in FL scenarios.

\begin{tcolorbox}[colback=blue!5!white,colframe=blue!75!black, boxsep=0mm, left=2mm, right=2mm, top=1mm, bottom=1mm]
\textbf{Highlight 17:} FedProx is more effective on image datasets, while MOON performs better with tabular datasets.
\end{tcolorbox}

\begin{table}[htbp]
    \centering
    \caption{The number of cases in which each aggregation algorithm achieved the best performance for each type of dataset} 
    \resizebox{\columnwidth}{!}{
    \begin{tabular}{||c|c|c|c|c|c|c||}
    \hline
    Dataset type &  \#Cases & FedAvg & Rand & FedProx & POC & MOON\\
    \hline\hline
    Image &  39 & 8 & 3 & 19 & 9 & 0 \\ \cline{1  - 7}
    Tabular & 49 & 1 & 0 & 7 &  9 & 32 \\ \cline{1  - 7}
    \cline{1  - 7}
    \multicolumn{2}{||c|}{Total best performance} & 9 & 3 & 26 & 18 & 32 \\ 
\cline{ 1 - 7}  
   
    \end{tabular}
    }
\label{tab:Bestperformance_dist_dataset}
\end{table}

Table~\ref{tab:Bestperformance_dist_dataset} examines the best-performing aggregation algorithms from the perspective of the dataset type used for training. It shows that FedProx outperforms all other algorithms on image datasets in ninetheen out of thirty-nine cases. In comparison, MOON generally surpasses other algorithms on tabular datasets in thirty-two out of forty-nine cases.

The effectiveness of FedProx on image datasets and MOON on tabular datasets can be attributed to their distinct optimization strategies. FedProx mitigates client drift by stabilizing updates, which is particularly beneficial for complex, high-dimensional image data. In contrast, MOON's contrastive learning framework enhances feature representation, making it more suited for tabular data, where feature relationships play a critical role. These differences explain their varying performance across dataset types.

%% file: 10-section-future-work.tex
We provide some design insights and opportunities, intending to help researchers
direct their efforts toward solving the effects of \hl{non-IID data}.

\paragraph{Quantifying \hl{the level of non-IID data}.}
Several works claim that the \hl{non-IID data} affects the performance
of FL models~\cite{lu2024federated, ma2022state, jamali2022federated}.
Nevertheless, for the first time, we demonstrate that the effect of the
\hl{non-IID data} is not the same under all the levels of heterogeneity (see
Figure~\ref{fig:all_accuracy}). 

Therefore, it is vital to quantify the \hl{non-IID data} level in FL. This
work uses the \textsf{HD} metric to measure the \hl{level of non-IID data}. However, we
encourage researchers to test different metrics, such as
JSD~\cite{nielsen2019jensen}, EMD~\cite{davis2023earth}, and Total Variation
distance~\cite{bhattacharyya2022approximating}, among others.

\paragraph{More effective methods to tackle high \hl{non-IID data}.}
This work showcases how the state-of-the-art methods to tackle \hl{non-IID data} (Rand, POC, FedProx, MOON) perform against FedAvg. The conclusion is that no algorithm works better than FedAvg in all the scenarios. Moreover, the better methods do not greatly improve FedAvg under high non-IID scenarios, as their gain is at most two percentage points. \hl{FedAvg remains competitive due to its computational efficiency and adequacy for moderately non-IID data, where its simplicity outperforms complex methods like FedProx or MOON that incur tuning overheads. However, in highly non-IID scenarios, adaptive approaches are essential, revealing a core trade-off between simplicity and adaptability. Thus, optimal algorithm selection depends on the specific non-IID data and system constraints in a given FL deployment.}

This phenomenon has also been studied and
claimed in the scarce work of empirical analysis of non-IID data and
methodologies~\cite{vahidian2023rethinking, abdelmoniem2022empirical,
mora2022federated, li2022federated}. Therefore, creating methods to alleviate the effect of high \hl{levels of non-IID data} appropriately is
needed to evolve and preserve FL. This aligns with the open
problems reported by Kairouz et al.~\cite{kairouz2021advances}.

\paragraph{Focusing on highly unbalanced data.}
In our experiments, we claim that the more significant decrease in
performance comparing CL and FL occurs in unbalanced datasets since it
relates to the challenge of learning from highly skewed and less
representative data (see Table~\ref{tab:FL_all_agg}). 
 
Thus, it is relevant to create solutions to tackle \hl{non-IID data} by considering the degree of unbalancedness that the labels might have
across the clients.

\paragraph{Studying spatiotemporal skew.}
At the time of writing this work, no analyses or empirical studies about the effect of the spatiotemporal skew on the performance of FL models exist. Thus, for the first time, we produce experiments to understand how different spatiotemporal \hl{non-IID data} levels affect an FL model's prediction power. The results show (see Table~\ref{tab:FL_all_agg_spatiotemporal_skew}) that high levels of space or time skews decrease the performance of the models, more specifically when the \textsf{HD} is higher than 0.75 (severe \hl{degree of non-IID data}). 

Thus, researchers may benchmark techniques to deal with space and time
skew in FL~\cite{shen2024decentralized, fu2023spatiotemporal,
zhou2022stfl} to determine the behavior under high \hl{non-IID data}
levels.

\paragraph{Methods to compare mixed \hl{non-IID data} types.}
Current tools and methods for synthetic partitioning centralized data into federated data~\cite{jimenez2024fedartml, zeng2023fedlab,lai2022fedscale, ogier2022flamby,percent_non_iid_method} focus on simulating one type of \hl{non-IID data} (label, feature, quantity, spatiotemporal skewness). Nevertheless, a more realistic scenario would be combining two or more types of non-IID data to evaluate the extent to which such mixes can alter the performance of FL models. Therefore, for research in FL purposes, it would be interesting to create methods to partition centralized data into federated clients that permit the control of \hl{non-IID data} level for two or more data skews simultaneously.

%% file: 11-section-conclusion.tex

This study provides a comprehensive empirical analysis of the non-IID effect in FL. Under controlled conditions, we benchmarked five state-of-the-art strategies for addressing non-IID data distributions, including label, feature, quantity, and spatiotemporal skew, placing particular focus on the relatively unexplored spatiotemporal dimension. We aim to standardize the methodology for studying \hl{non-IID data} in FL by using \textsf{HD} to quantify data distribution differences. Our findings reveal the significant impact of labels and spatiotemporal skews of non-IID types on FL model performance. We also demonstrate that the model’s performance drop appears at a double threshold. When \textsf{HD} is higher than 0.5 and 0.75, higher damage and a steeper decrease in performance slope occur. Moreover, our results suggest that the FL performance is heavily affected, mainly when the \hl{degree of non-IID data} is extreme. Thus, we offer valuable recommendations for researchers to address \hl{non-IID data}. This work represents the most thorough examination of \hl{non-IID data} in FL to date, providing a robust foundation for future research in FL.